\documentclass[review,5p,twocolumn]{elsarticle}

\usepackage{amssymb}
\usepackage{algorithm, algorithmic}
\usepackage{graphicx}
\usepackage{subfigure}
\usepackage{tabularx}
\usepackage{setspace}

\usepackage{epstopdf}
\usepackage{amsmath}

\usepackage{longtable}
\usepackage{threeparttable}
\usepackage{multirow}
\usepackage{color}

\usepackage{caption}
\usepackage{float}
\usepackage{multicol}
\usepackage{amsmath}
\usepackage{autobreak}
\usepackage{hyperref}
\usepackage{CJKutf8}

\journal{Knowledge-Based Systems}

\begin{document}
\begin{CJK}{UTF8}{gbsn}
\begin{frontmatter}



% \title{OpenClinicalAI: Enabling AI to Diagnose Alzheimer's Disease in real-world clinical setting}
\title{OpenClinicalAI: An Open and Dynamic Model for Alzheimer's Disease Diagnosis}

\author[label1,label2]{Yunyou Huang}
\author[label1,label2]{Xiaoshuang Liang}
\author[label1,label2]{Xiangjiang Lu}
\author[label1,label2]{Xiuxia Miao}
\author[label1,label2]{Jiyue Xie}
\author[label1,label2]{Wenjing Liu}
\author[label4]{Fan Zhang}
\author[label4]{Guoxin Kang}
\author[label3]{Li Ma}
\author[label1,label2]{Suqin Tang}
\author[label6]{Zhifei Zhang\corref{cor1}}
\ead{zhifeiz@ccmu.edu.cn}
\author[label4,label5]{Jianfeng Zhan\corref{cor1}}
\ead{zhanjianfeng@ict.ac.cn}
\cortext[cor1]{Corresponding author:}

\affiliation[label1]{organization={Key Lab of Education Blockchain and Intelligent Technology, Ministry of Education, Guangxi Normal University},%Department and Organization
            city={Guilin},
            country={China}}

\affiliation[label2]{organization={Guangxi Key Lab of Multi-Source Information Mining and Security, Guangxi Normal University},%Department and Organization
            city={Guilin},
            country={China}}

%\affiliation[label2]{organization={State Key Laboratory of Computer Architecture, Institute of Computing Technology, Chinese Academy of Sciences},%Department and Organization
%            city={Beijing},
%            country={China}}

\affiliation[label4]{organization={University of Chinese Academy of Sciences},%Department and Organization
            country={China}}

\affiliation[label3]{organization={Guilin Medical University},%Department and Organization
            city={Guilin},
            country={China}}

\affiliation[label6]{organization={Department of Physiology and Pathophysiology, Capital Medical University},%Department and Organization
            country={China}}
            
%Department of Physiology and Pathophysiology, Capital Medical University, Beijing, 100069, China.
%multi-task learning and deep reinforcement learning

\affiliation[label5]{organization={International Open Benchmark Council}}

\begin{abstract}
Although Alzheimer's disease (AD) cannot be reversed or cured, timely diagnosis can significantly reduce the burden of treatment and care. Current research on AD diagnosis models usually regards the diagnosis task as a typical classification task with two primary assumptions: 1) All target categories are known a priori; 2) The diagnostic strategy for each patient is consistent, that is, the number and type of model input data for each patient are the same. However, real-world clinical settings are open, with complexity and uncertainty in terms of both subjects and the resources of the medical institutions.  This means that diagnostic models may encounter unseen disease categories and need to dynamically develop diagnostic strategies based on the subject's specific circumstances and available medical resources. Thus, the AD diagnosis task is tangled and coupled with the diagnosis strategy formulation. To promote the application of diagnostic systems in real-world clinical settings, we propose  OpenClinicalAI for direct AD diagnosis in complex and uncertain clinical settings. This is the first powerful end-to-end model to dynamically formulate diagnostic strategies and provide diagnostic results based on the subject's conditions and available medical resources. OpenClinicalAI combines reciprocally coupled deep multiaction reinforcement learning (DMARL) for diagnostic strategy formulation and multicenter meta-learning (MCML) for open-set recognition. The experimental results show that OpenClinicalAI achieves better performance and fewer clinical examinations than the state-of-the-art model. Our method  provides an opportunity to embed the AD diagnostic system into the current health care system to cooperate with clinicians to improve current health care.
\end{abstract}


\end{frontmatter}


\section{Introduction}
\label{sec:intro}
Alzheimer's disease is an incurable disease that heavily burdens our society (the total cost of caring  for individuals with AD or other dementias is estimated at $\$321$ billion)~\cite{https://doi.org/10.1002/alz.12638,hebert2001annual,hebert2013alzheimer,alzheimer20182018,frigerio2019major}. Early and accurate diagnosis of AD is crucial for effectively managing the disease and has the potential to save nearly $\$7$ trillion in medical and care costs.~\cite{https://doi.org/10.1002/alz.12638,alzheimer20182018}. However, it is estimated that $28$ million of the $36$ million people with dementia worldwide have not received a timely and accurate diagnosis due to limited medical resources, availability of experts, etc.~\cite{prince2018world}. Artificial intelligence (AI), as one of the  technologies with the most potential to improve medical services, is widely employed in AD diagnosis research~\cite{tanveer2020machine,mahajan2020machine}. Daniel et al.~\cite{stamate2019metabolite} utilized plasma metabolites as inputs for Extreme Gradient Boosting (XGBoost) and demonstrated their potential in diagnosing AD. Xin et al.~\cite{xing2020dynamic} proposed an approximate rank pooling method to transform $3$D nuclear magnetic resonance images (MRI) into $2$D images, followed by the utilization of a $2$D convolutional neural network (CNN) for AD classification. Qiu et al.~\cite{qiu2022multimodal} reported a multimodal diagnosis framework that used CNN to capture the critical features of MRI images and then used Categorical Boosting (CatBoost) to detect the presence of AD and cognitively normal (CN) from demographics, medical history, neuroimaging, neuropsychological testing, and functional assessments.

\begin{figure*}
\centering
\begin{minipage}[a]{0.48\textwidth}
%\caption*{a.}
\leftline{\textbf{a}}
\includegraphics[height=7.5cm, width=5cm]{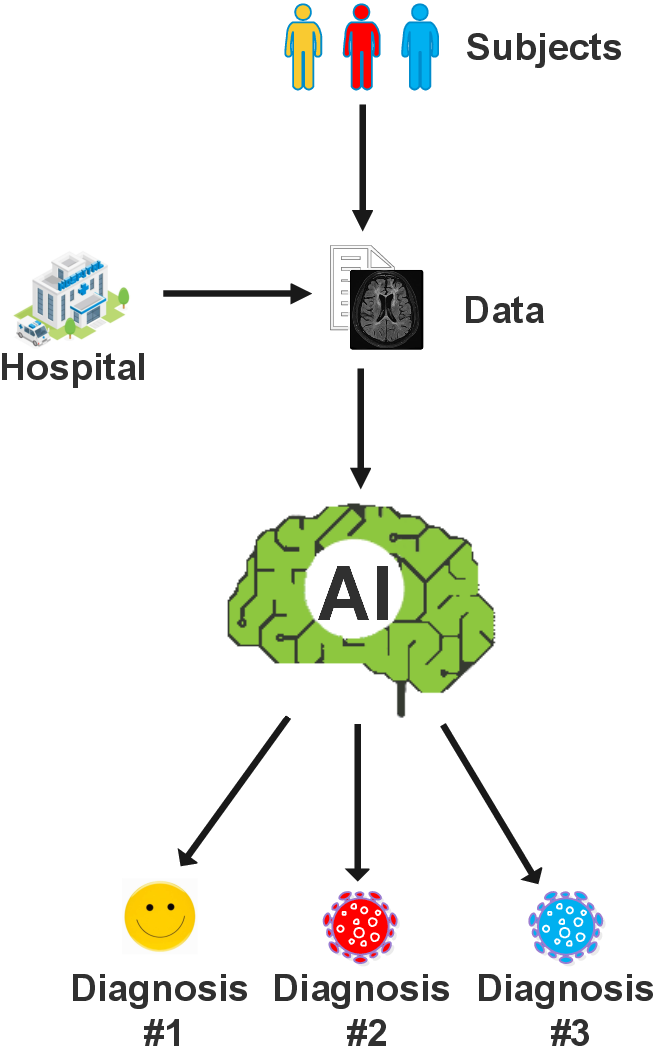} 
\end{minipage}
%\hspace{0.5cm}
\begin{minipage}[a]{0.48\textwidth}
\leftline{\textbf{b}}
\includegraphics[height=8cm, width=9cm]{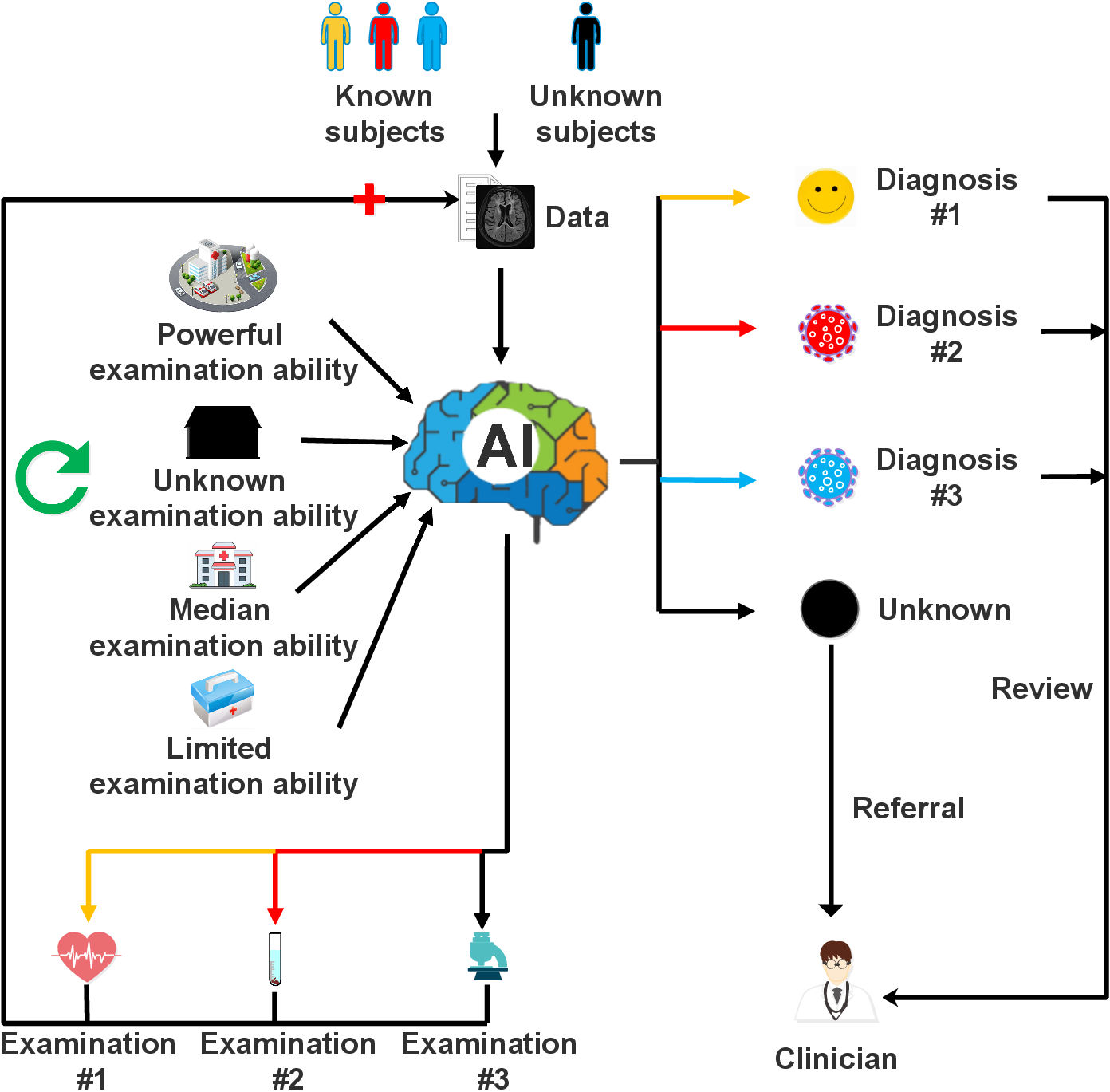} 
\end{minipage}\\
\quad
\caption{\textbf{The workflow of the baseline clinical AI system and OpenClinicalAI. a,} The workflow of the mainstream AI-based diagnostic systems for closed settings. \textbf{b,} The workflow of models in the real-world setting.
\label{fig1}}
\end{figure*}

As Fig. ~\ref{fig1} (a) shows, the AD diagnosis models in previous works are designed for the closed clinical setting with the following primary assumptions: (1) all categories of subjects are known a priori (subject categories are aligned in training and test sets by inclusion and exclusion criteria); (2) the same diagnostic strategies are used to diagnose all subjects (Delete subjects with missing data of a certain type or fill in missing data for subjects in data preprocessing); and (3) all medical institutions are capable of executing the prescribed diagnostic strategies~\cite{qiu2020development,xing2020dynamic,bohle2019layer}. This makes the AD diagnosis task an independent typical classification task. However, as Fig.~\ref{fig1} (b) shows, the real-world clinical setting is open with uncertainty and complexity: (1) The categories of subjects in real-world clinical settings are not all known in advance and may include unknown categories that do not appear during the development of AI models. This implies that the test set may contain subject categories not present in the training set. This transforms AD diagnosis into an open-set recognition problem instead of it  being a conventional classification problem. (2) Each subject is unique, and there is no one-size-fits-all diagnosis strategy. This  results in variations in the amount and types of data across different subjects. (3) The conditions of medical institutions vary and are not known in advance; e.g., positron emission tomography (PET) is available in some hospitals but most hospitals in underdeveloped areas are not equipped with PET. Thus, the AD diagnosis task is an open-set recognition task tangled and coupled with the formulation of diagnosis strategies.

Currently, to tackle the challenges in the real-world clinical setting, open-set recognition technology has emerged in various fields~\cite{geng2020recent}. However, open-set recognition only considers unknown categories while overlooking the complexities of subjects and the constraints posed by medical resources. In the real-world clinical setting,  this  hinders the application of open-set recognition technology to AD diagnosis. %Note, automatic diagnosis 
Therefore, the critical problem is how to simultaneously handle the uncertainty and complexity of subjects and medical resources in AD diagnosis within the real-world clinical setting.

\begin{figure*}[ht]
\centering
\includegraphics[height=8cm, width=13cm]{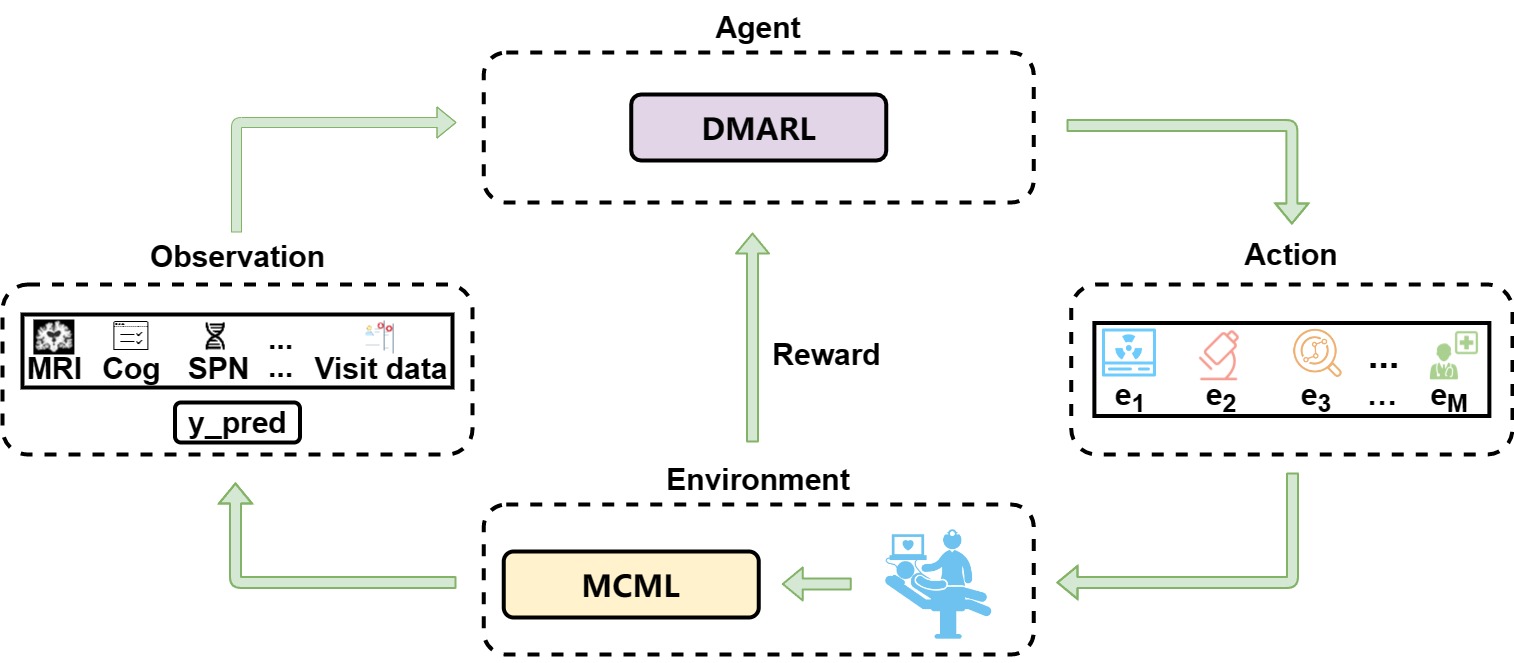}
\caption{\textbf{The reinforcement learning framework based on AD diagnosis and inspection recommendation model.} \label{RML_first}}
\end{figure*}

In this paper, we explore the AD diagnosis task from a new perspective of both subjects and medical institutions and redefine AD diagnosis as an open, dynamic real-world clinical setting recognition problem. Specifically, clinicians first formulate a preliminary diagnosis strategy based on the individual's condition and available medical resources after enquiring about the basic information involving the subject. Second, the subjects undergo examinations in accordance with the diagnostic strategy.
Third, the model merges all available information, categorizes subjects into prespecified or unknown categories, or adjusts the diagnostic strategy and returns to the second step. To 
%Quality Control Editor: Please ensure that the intended meaning has been maintained in the following edit.
realize this approach, 
as illustrated in Fig.~\ref{RML_first}, we propose OpenClinicalAI, an open and dynamic deep learning model to directly diagnose subjects in complex and uncertain real-world clinical settings. OpenClinicalAI comprises two tangled and coupled modules: deep multiaction reinforcement learning (DMARL) and multicenter meta-learning (MCML). MCML utilizes AutoEncoder and meta-learning techniques to diagnose subjects based on the subject's data obtained from the diagnostic strategy formulated by DMARL. It serves as an environmental simulator, providing feedback to DMARL. 
DMARL is a multitask learning model that functions as an agent which  dynamically adjusts the subject's diagnosis strategy and 
%Quality Control Editor: Please ensure that the intended meaning has been maintained in the following edit.

acquires fresh
 examination data based on rewards from MCML and the subject's current examination data.

To summarize, our contributions are as follows:

 \begin{itemize}

\item[$\bullet$] OpenClinicalAI is the first end-to-end coupling model designed specifically for direct AD diagnosis in complex and uncertain real-world clinical settings. It dynamically develops $35$ different diagnosis strategies according to different subject situations and $40$ different examination abilities of medical institutions on the test set. The framework of OpenClinicalAI is domain-independent and can be extended to other diseases to promote the development of real-world clinical setting diagnosis.

\item[$\bullet$] The newly designed DMARL enables OpenClinicalAI to dynamically adjust the diagnosis strategy according to the subject's current data, and MCML provides feedback on classification information and rewards. Its novelty lies in designing a multitask model for selecting the next clinical examinations for a subject and avoiding a step-by-step form of reinforcement learning.

\item[$\bullet$] The novel MCML promotes open-set recognition of AD based on the diagnosis strategy from DMARL and provides disease classification information and rewards for DMARL. It uses AutoEncoders to retain more general features for open set recognition and uses clustering to divide subjects of the same category into multiple subcategories. This  improves the accuracy of meta-learning algorithms in calculating subject similarity, thereby providing unknown subject recognition. In addition, it is also used as an environmental simulator to evaluate the rewards of the diagnostic strategies formulated by DMARL, and the disease classification information is used as the input of DMARL to help dynamically formulate diagnostic strategies.

\item[$\bullet$] In the closed clinical setting, OpenClinicalAI shows a comparable performance to the current state-of-the-art model in terms of the AUC (area under the receiver operating characteristic curve) metric. At the same time, less than $10\%$ of subjects are required to have an MRI image. In the real-world clinical setting, our model outperforms the current state-of-the-art model by: (1) an absolute increase of $11.02\%$ and $11.48\%$ (AD diagnosis and cognitively normal diagnosis) in AUC and 30.09\% and $47.94\%$ in sensitivity; and (2) an absolute reduction of $68.93\%$ MRI for subjects. In addition, it has $93.96\%$ in sensitivity for identifying unknown categories of subjects.
\end{itemize}

\section{Related Work}
\label{sec:Relatework}

\textbf{AD diagnosis model.} With progress in machine learning and deep learning, AD diagnostic models have gained significant attention in the medical community. AD diagnosis studies can be divided into nonimage-based, image-based, and multimodal-based studies. Aljovic et al.~\cite{aljovic2016artificial} designed a linear feed-forward (FF) neural network that used biomarkers (albumin ratio, AP$40$, AP$42$, tau-total, and tau-phospho) as input to classify Alzheimer's disease. This work proved that it is possible to use biomarker data for AD discrimination. Lu et al.~\cite{lu2022practical} trained a $3$-dimensional sex classifier Inception-ResNet-V$2$ as a base model in transfer learning for AD diagnosis. They then made it suitable for $3$-dimensional MRI inputs and  achieved high accuracy in a large collection of brain MRI samples ($85,721$ scans from $50,876$ participants). Qiu et al.~\cite{qiu2020development} proposed an interpretable multimodal deep learning framework that used a fully convolutional network (FCN) to perform MRI image feature extraction and then used a traditional multilayer perceptron (MLP) with multimodal inputs (image and text features) to classify the disease and generate disease probability maps. Many studies have shown that the performance of multimodal models is better than that of single-modal models because different modal information can complement each other to help diagnose AD.

\textbf{Open-set recognition.} 
% To address the real-world clinical setting problem, many open-set recognition(OSR) technologies are proposed, and are able to divide into discriminative model and generative model~\cite{geng2020recent}.
Various open-set recognition (OSR) techniques have been proposed to solve the problem of unknown categories in the real world. They can be divided into discriminant and generative models~\cite{geng2020recent}. Scheirer et al.~\cite{scheirer2012toward} proposed a preliminary solution, a $1$-vs.-Set machine, which sculpts a decision space from the marginal distances of a 1-class or binary SVM with a linear kernel. Bendale et al.~\cite{bendale2016towards} introduced a  model called layer-OpenMax, which estimates the probability of an input being from an unknown class to adapt deep networks for open-set recognition. Oza et al.~\cite{oza2019c2ae} proposed a deep neural network-based model---C$2$AE, which uses class-conditioned autoencoders with novel training and testing methodologies. Yang et al.~\cite{yang2019open} proposed a model based on a generative adversarial network (GAN) to address open-set human activity recognition without manual intervention during the training process. Geng et al.~\cite{geng2020collective} proposed a collective decision-based OSR framework (CD-OSR) by slightly modifying the hierarchical Dirichlet process (HDP). This method aimed to extend existing OSR for new class discovery while considering correlations among the testing instances. Perera et al.~\cite{perera2020generative} utilized self-supervision and augmented the input image to learn richer features to improve separation between classes. These methods tend to use powerful generative models to mimic novel patterns.

\textbf{Deep reinforcement learning.}
Deep learning is already widely applied in the medical field for prognosis prediction and treatment recommendation. Saboo et al.~\cite{NEURIPS2021_af1c25e8} simulated the progression of Alzheimer's disease (AD) by integrating differential equations (DEs) and reinforcement learning (RL) with domain knowledge. DEs serve as an emulator, providing the relationships between some (but not all) factors associated with AD. The trained RL model acts as a proxy to extract the missing relationship by optimizing the objective function over time. The model uses baseline ($0$ year) characteristics from aggregated and real data to personalize a patient's $10$-year AD progression. Quan et al.~\cite{Zhang_2021} proposed an interpretable deep reinforcement learning model to reconstruct compressed sensing MRI images. Komorowski et al.~\cite{komorowski2018artificial} developed an artificial intelligence (AI) clinician model using a reinforcement learning agent. This model extracts implicit knowledge from patient data based on the lifetime experience of human clinicians and analyzes numerous treatment decisions, including mostly suboptimal decisions, to learn optimal treatment strategies. Experiments have shown that, on average, the value of treatments selected by the AI clinician model is reliably higher than that of human clinicians.

Deep reinforcement learning (DRL) is a combination of deep learning and reinforcement learning algorithms that learn and optimize decision-making strategies through interactions with the environment. However, current research often treats disease diagnosis as a one-step task, where the model inputs patient information and obtains a classification. This approach has led to a neglect of the potential application of DRL in diagnostic tasks.

\section{Problem Formulation}

In this work, the diagnosis of AD involves dynamically developing diagnostic strategies based on the subject's situation and the medical institution, ultimately determining whether the subject belongs to the unknown category. To address the dynamic interaction process between subjects and medical institutions, we propose a reinforcement learning model, namely, OpenClinicalAI, to solve this problem. The detailed definition of the AD diagnosis problem is as follows:
\par
\textbf{Agent:} The agent is capable of perceiving the state of the external environment and the rewards and uses this information to learn and make decisions. In this problem, we employ the DMARL model as the agent, which takes observation of the environment as input and formulates recommended next clinical examinations based on the subject's conditions and the medical institution.\par
\textbf{Environment:} The environment refers to all objects external to the intelligent agent, whose observation is influenced by the actions of the agent and provides corresponding rewards to the agent. In our problem, the environment encompasses three components: the MCML model, subjects, and medical institutions. Medical institutions conduct clinical examinations on subjects according to the action suggested by the agent. The data generated by these clinical examinations serve as input for the MCML model, which generates intermediate diagnostic results. The environment provides feedback to the intelligent agent using the diagnostic intermediate results and the latest clinical data while simultaneously calculating the corresponding reward.\par
\textbf{Observation:} The agent's observation of the environment include the intermediate diagnostic results from MCML and the currently available clinical data of the subjects.\par
\textbf{Action}: The agent interacts with the environment via actions where the  actions include $12$ types of clinical examinations. An  action may consist of one or more clinical examinations.\par
\textbf{Reward:} The reward is the bonus that an agent gets once it takes an action in the environment at the given time step $t$. In this paper, the reward is measured as the degree of improvement in the probability of intermediate diagnostic results obtained by the MCML model after taking an action.\par
\textbf{Discount factor ($\gamma$)}: The discount factor measures the importance of future rewards to the agent in the current state. In this paper, $\gamma$ represents the set of hyperparameters of the OpenClinicalAI model. During the gradient descent process, the model will adjust these hyperparameters to obtain the maximum expected reward.\par

We use $E =\{e_1, e_2, \cdots, e_{13}\}$ to represent $13$ clinical examinations performed for Alzheimer's disease, $D =\{d_1, d_2, \cdots, d_{13}\}$ to represent the data obtained by a subject undergoing the corresponding clinical examination, and $T =\{t_0, t_1, \cdots, t_l\}$ to represent the time step series of the model, where $l$ is determined by the model according to the conditions of the subject and the medical institution. Using $S=\{s_i\}_{i=1}^n$ to represent the data for $n$ subjects contained in the dataset, where $s_{t_l}^{(i)} = \{d_k\}^m$ represents the $m$ clinical examination data that $subject\_i$ has up to the $t_l$ time step with  $k \in [1, 13]$, $m \in [1, 13]$, $len(\{d_k\}^m)=m$, $\{d_k\} ^m\subseteq D$. Using $s_{t_0}^{(i)}$ represents the original data of $subject\_i$. $a_{t_l}=\{e_j\}^u$ indicates the next clinical examination recommended by the model (which contains $u$ clinical examinations) at the $t_l$ time step, where $j \in [2, 13]$, $u \in [1, 12]$, $len(\{e_j\}^u)=u$, $\{e_j\} ^u\subseteq E$. $da_{t_l}=\{d_j\}^u$ is used to represent the clinical examination data obtained by the subject after executing $a_{t_l}$. Then, update the $subject\_i$ data $s_{t_{l + 1}}^{(i)} = s_{t_l}^{(i)} \cup 
da_{t_l}^{(i)}$.
\par

In particular, we call the set of actions selected by the model in $t_1$ to $t_l$ time steps a diagnostic strategy $ds=\{a_{t_1} \cup a_{t_2} \cup \cdots \cup a_{t_l}\}$, and the set of diagnostic strategies is represented by $DS=\{ds_q | ds_q = \{e_j\}^h, h \leq m\}, |DS|=Q$, where $Q$ represents the number of diagnostic strategies generated based on the combination of $m$ clinical examinations contained by the subject data, which is determined by the model based on the subject and the medical institution. Therefore, we use $s^{(i)}_{ds_q}$ to represent the data obtained by $subject\_i$ after executing the diagnostic strategy $ds_q$. In addition, $s^{(i)}_{ds_q}$ as an input of the MCML model will obtain the corresponding diagnosis $y^{(i)}_{pred\_ds_q}$, and $s_{t_l}^{(i)}$ as an input of the MCML model will obtain the corresponding intermediate diagnosis $y^{(i)}_{pred\_s_{t_l}}$. Thus, the observation of the $lth$ time step is $\{s_{t_l}^{(i)}, y^{(i)}_{pred\_s_{t_l}}\}$, and the reward after executing action $a^{(i)}_{t_l}$ is $r_{t_l}^{(i)}$.

Based on the aforementioned definitions, $Trajectory (\tau)$ is a sequence of states, actions, and rewards that are generated by the model's interaction with $subject\_i$ over a time step of duration $l:\tau = \{((s_{t_1}^{(i)}, y_{pred\_t_1}^{(i)}), a_{t_1}^{(i)},r_{t_1}^{(i)}),((s_{t_2}^{(i)}, y_{pred\_t_2}^{(i)}), a_{t_2}^{(i)},r_{t_2}^{(i)}), \cdots,((s_{t_l}^{(i)}, \\y_{pred\_t_l}^{(i)}), a_{t_l}^{(i)},r_{t_l}^{(i)})\} $

\section{Methodology}
\label{sec:Materials and methods}

\begin{figure*}[h]
\centering
\includegraphics[height=16cm, width=15cm]{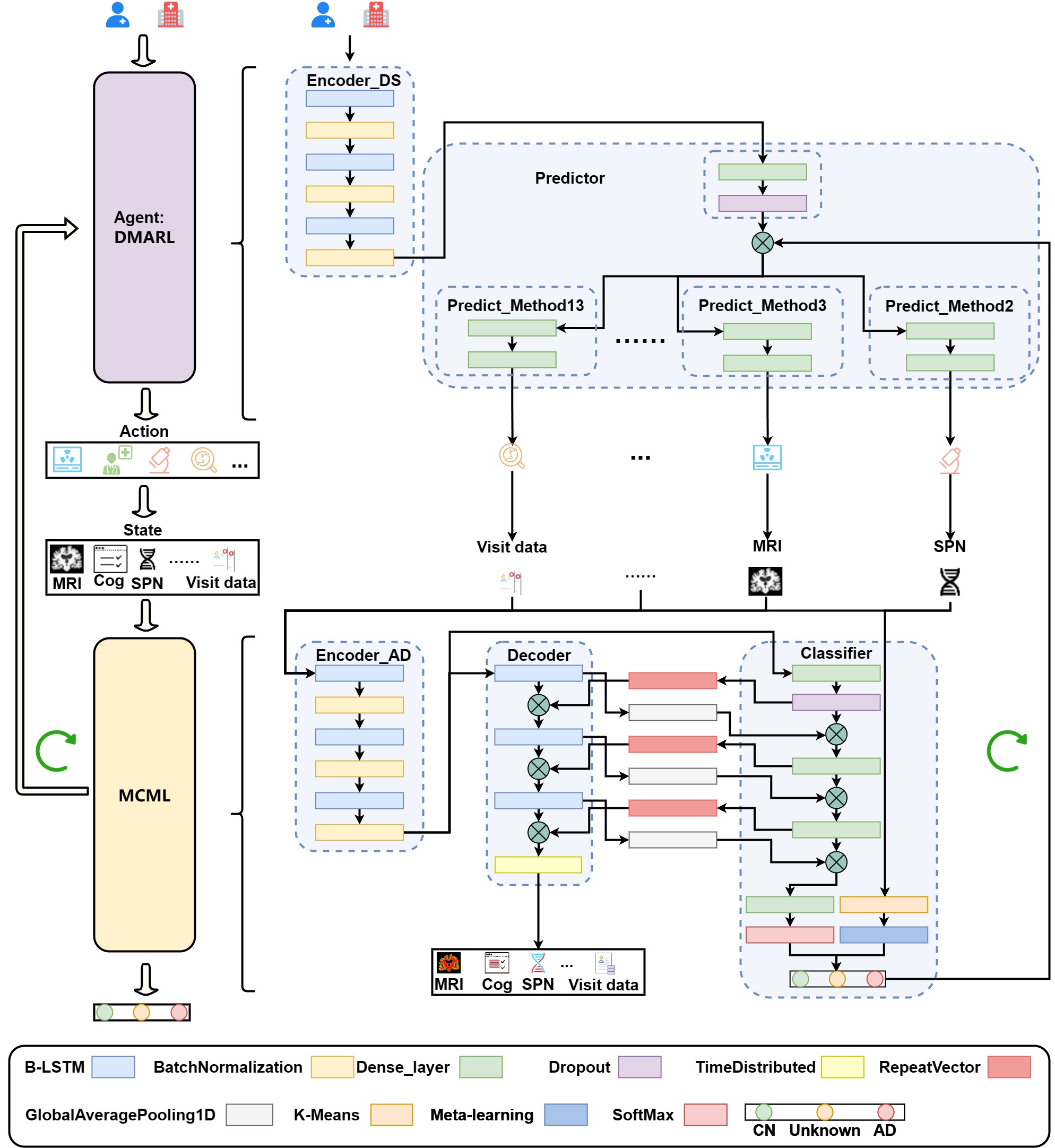}
\caption{\textbf{The framework detail of OpenClinicalAI.} \label{RML_Detail}}
\end{figure*}

\subsection{Data preparation}

For each subject, if there is  more than one visit, each visit is considered to be an independent sample. Multiple categories of data are generated at each visit based on the diagnostic strategy. This  results in each sample usually containing multiple types of data. As for medical images, we first convert the data from DICOM format to NIFTI format using the dcm$2$nii library. Then, we perform image registration using the ANTs library~\cite{5cf2edbbe75b4f90ac4c89e6ef1bc3f6,smith2004advances,senecaimproved}. Next, we convert the $3$D image into $2$D slices and transform the image from grayscale to RGB. Finally, we utilize a pretrained model called DenseNet$201$ to extract features from the $2$D slices~\cite{huang2017densely}. For genetic data, we extract $70$ single nucleotide polymorphisms (SNPs) that are highly related to AD and represent each SNP using a one-hot encoding method~\cite{lambert2013meta,kunkle2019genetic,desikan2017genetic}. To accommodate the varying dimensions of each data category, the number of data categories included in each visit, and the number of past visits for each subject, we use a unified data representation framework, as illustrated in Supplementary Figure. S1. In this framework, we use an array with a shape of $1\times2090$ to represent the examination category in the subject's visit. The shape of our data is $n\times2090$, where n represents the number of data categories for the subject.

\subsection{OpenclinicalAI}
As shown in Fig.~\ref{RML_Detail}, OpenclinicalAI is based on the status of subjects and medical institutions to directly formulate diagnostic strategies and provide accurate diagnostic results.  This approach enables the system to effectively handle the inherent uncertainty and complexity of real-world clinical settings. This model is composed of two coupled elements: (1) DMARL (refer to Section~\ref{dmarl}) combines encoder, multitask learning, and deep reinforcement learning to dynamically adapt the subject's diagnostic strategy based on the specific circumstances of the hospital, subject, and feedback from MCML. (2) MCML (refer to Section~\ref{mcml}) takes the data generated by DMARL's predictions as input, employs AutoEncoder to extract more general features, calculates more precise sample-to-class center distances, and integrates meta-learning techniques to facilitate AD identification in open clinical environments. In summary, the objective of OpenclinicalAI can be formulated as follows:

\begin{equation}
\begin{aligned}
  \label{real_world1}
%   \begin{split}
  &L_{r}(W)=\sum_{i=1}^n{[\alpha l1(\hat{y_i}, y_{i})} +\beta l2(W)+\lambda l3(X_i,W) \\& \qquad \qquad+\mu l4(X_i,\hat{X_i})] 
\end{aligned} 
\end{equation}

where $\hat{y_i}$ and $y_i$ indicate the predicted class probability and true label of the subject, respectively. Similarly, $X_i$ and $\hat{X_i}$ denote the subject's original input features and reconstructed features, respectively. $\alpha$, $\beta$, $\lambda$, and $\mu$ correspond to the hyperparameters of each loss, while $W$ signifies the weight of the network. Specifically, $l1$ is the softmax cross-entropy function, $l2$ represents the $L2$ regularized loss function, $l3$ utilizes uncertainty to weigh losses in a multitask learning scenario, and $l4$ captures the reconstruction loss of $X_i$.

\subsubsection{MCML for AD recognition and environment simulation}
\label{mcml}
The architecture of MCML is inspired by Generative Discriminative Feature Representations~\cite{perera2020generative} and meta-learning-based OpenMax~\cite{bendale2016towards}. As depicted in Fig.~\ref{RML_Detail}, MCML combines the advantages of the previous two works. In addition, (1) MCML further improves the retention of generalized features by fostering information interaction between each layer of the decoder and classifier. This mitigates the risk of the classifier solely focusing on the most relevant features for classification. (2) MCML utilizes k-means clustering to partition subjects of the same category into multiple subtypes. It then calculates the distance between subjects and the center of the nearest subtype. This approach reduces the likelihood of atypical subjects being misclassified as unknown categories due to their substantial distance from the category center. Specifically, MCML comprises three components: $Encoder\_AD$, $Decoder$, and $Classifier$. These components classify subjects into AD, CN, and unknown categories based on specific diagnostic strategies. The $Decoder$ is structured as a mirror image of the $Encoder\_AD$ to enable the model to capture more subject-specific characteristics associated with AD and CN classification. The $Classifier$ consists of three dense layers and a SoftMax/OpenMax layer which facilitates subject classification in the open clinical setting. In summary, the loss function for the MCML module is formulated as follows:

\begin{equation}
\left\{
\begin{aligned}
  \label{real_world2}
%   \begin{split}
  &{l1(\hat{y_i}, y_{i})} = -[y_{i}log(p_{i})+(1-y_{i})log(1-p_{i})]\\
  &{l4(X_i,\hat{X_i})} = ||X_i-\hat{X_i}||_{2}^2\\
    &\hat{y_i}=f(W(X_i))\\
  &\hat{y_{i\_un}}=1-\sum_{j=1}^2\hat{y_i}[j]
%   \end{split}
\end{aligned}
\right.    
\end{equation}
where $p_{i}$ denotes the probability assigned to the subject's cognitive state prediction, and the cross entropy loss $l1$ is employed to quantify the disparity between the subject's current cognitive state, determined after an examination, and the cognitive state predicted by the MCML model. The score modifier $f$ is based on $EVT$~\cite{bendale2016towards}, and $\hat{y_{i\_un}}$ represents the probability that the sample belongs to an unknown category.

\subsubsection{DMARL for diagnostic strategy development}
\label{dmarl}

%\textcolor{red}{
As depicted in Fig.~\ref{RML_Detail}, the DMARL module consists of an encoder, $Encoder\_DS$, and a predictor, $Predictor$. As an agent, DMARL takes the observation of the subject and the feedback of MCML as inputs. To effectively incorporate variable-length clinical data based on different diagnostic strategies, we employ a bidirectional long short-term memory (B-LSTM) network in the encoding pathway. $Encoder\_DS$ comprises three layers of B-LSTM units. The primary objective of DMARL is to utilize existing medical resources, subject observations, and the classification confidence from MCML to predict the subsequent clinical examinations for each subject and dynamically formulate an optimal diagnostic strategy. To integrate the output of $Encoder\_DS$ with the classification confidence from MCML, $Predictor$ consists of $13$ dense layers. Furthermore, to determine the appropriate clinical examination for each subject, the model incorporates $12$ independent sigmoid classifiers. In addition, due to the lack of labels for the next clinical examination, DMARL will be trained according to the rewards of MCML. Thus, the loss function for the DMARL module is expressed as follows:

\begin{equation}
\begin{aligned}
  \label{real_world3}
%   \begin{split}
  &l3(X_i,W) = \sum_{i=1}^{13}[\frac{1} {2\delta_{i}^2}l1(W)+log\delta_{i}] 
\end{aligned} 
\end{equation}

The sum of $l3$ losses is calculated by evaluating $13$ recommended subtasks ($e$) for examination, where $\delta_{i}$ is an observation noise scalar of the output of ith examination~\cite{kendall2018multi}.

\subsection{Reward of the diagnosis strategy}
Although there have been significant efforts to improve the interpretability and internal logic of deep learning, understanding the behavior of deep learning models remains challenging~\cite{zhang2020effect,linardatos2021explainable}. We do not know whether the diagnosis strategy of the AI model needs to be consistent with that of an human expert
. Therefore, in this work, we train the DMARL module based on the reward generated by MCML. The ultimate goal of OpenclinicalAI is to accurately identify AD patients using MCML. The subsequent examination for each subject is determined by whether it leads to a higher predicted probability for the correct category and lower predicted probabilities for other categories. The reward of MCML is calculated using Algorithm~\ref{alg1}.

\floatname{algorithm}{Algorithm}
\setcounter{algorithm}{0}

\begin{algorithm}[h]
%\scriptsize
         \renewcommand{\algorithmicrequire}{\textbf{Input:}}
         \renewcommand{\algorithmicensure}{\textbf{Output:}}
         \caption{\textbf{The calculation of reward.}\label{alg1}}
         %\label{alg1}
         \begin{algorithmic}[1]
                   \REQUIRE The label $y_{true}$ and the diagnosis strategy set $DS$ of a subject. The prediction set $Pred=\{y_{pred\_ds_q}, ds_q\in DS\}$ of the MCML.
                   \ENSURE The reward $r_{t_l}$ of the action $a_{t_l}$ set $OAR=\{<(s_{t_l} \cup y_{pred\_t_l}),a_{t_l},r_{t_l}>\}$
                   %Next examination set $next\_exam$
                   \STATE Sort the $DS$ by the number of clinical examinations in every diagnosis strategy.
                   \FOR{$ q=0  \quad to  \quad len(DS) $}
                   \FOR{$ v=q+1  \quad to  \quad len(DS) $}
                   \IF{$ds_{q} \subset ds_{v}$}
                   \STATE $s_{t_l}=s_{ds_q}$
                   \STATE $s_{t_{l+1}}=s_{ds_v}$
                   \STATE $y_{pred\_t_l}=y_{pred\_ds_q}$
                    \STATE  $r_{t_l}=sum(y_{true}\times y_{pred\_ds_v}-y_{true}\times y_{pred\_ds_q})+sum(\sim y_{true}\times y_{pred\_ds_q}-\sim y_{true}\times y_{pred\_ds_v})$
                \IF{$r_{t_l}>0$}
                \STATE $a_{t_l}=ds_{v}-ds_{q}$
                %\STATE $s^{(i)}_{T_{l+1}}= s^{(i)}_{T_{l}} \cup ad^{(i)}_{t_l} $
                \STATE Add $<(s_{t_l} \cup y_{pred\_t_l}),a_{t_l},r_{t_l}>$ into $OAR$
                \ENDIF
                    \ENDIF
                   \ENDFOR
                   \ENDFOR
                   \RETURN $OAR=\{<(s_{t_l} \cup y_{pred\_t_l}),a_{t_l},r_{t_l}>\}$
         \end{algorithmic}
\end{algorithm}

\subsection{Model training}
The  training process is divided into two stages. In the first stage, the procedure is as follows: (1) For each $subject\_i$, the diagnostic strategy set $DS$ is generated according to $m$ examinations contained in the data $s^{(i)}=\{d_k\}^m$. It should be noted that the types and number of clinical examinations are not the same between subjects since the diagnosis strategy will change dynamically according to different subjects and available medical resources. (2) Every $s^{(i)}_{ds_q} \subseteq S^{(i)}$ is considered to be an independent sample and forms the dataset $D_{diagnosis}=\{<s^{(i)}_{ds_q},Y^{(i)}>\}$, where $Y^{(i)}$ represents the true diagnostic category label of $subject\_i$. (3) The MCML model is trained and tested on $D_{diagnosis}$, and the intermediate diagnosis result $Pred=\{y_{pred\_ds_q}, ds_q\in DS\}$ is obtained. In the second stage, the process is as follows: (1) For every $(s^{(i)}, y_{pred}^{(i)})$, we obtain the reward $r^{(i)}$ and $a^{(i)}$ by Algorithm~\ref{alg1}, and form the dataset $D_{examination}=\{((s_{t_1}^{(i)}, y_{pred\_t_1}^{(i)}), a_{t_1}^{(i)},r_{t_1}^{(i)}),((s_{t_2}^{(i)}, y_{pred\_t_2}^{(i)}), a_{t_2}^{(i)},r_{t_2}^{(i)}), \cdots, ((\\ s_{t_l}^{(i)}, y_{pred\_t_l}^{(i)}), a_{t_lL}^{(i)},r_{t_l}^{(i)})\}$. (2) We then train and test the DMARL on $D_{examination}$.

\subsection{Prediction}

To deliver  the final diagnosis result, it is imperative to dynamically adjust the diagnosis strategy. The prediction process is delineated in Algorithm~\ref{alg_3}.

\begin{algorithm}[h]
%\scriptsize
         \renewcommand{\algorithmicrequire}{\textbf{Input:}}
         \renewcommand{\algorithmicensure}{\textbf{Output:}}
         \caption{\textbf{The prediction algorithm.}\label{alg_3}}
         %\label{alg1}
         \begin{algorithmic}[1]
                   \REQUIRE The base information $data_{base}$ and history recodes $data_h$ for a subject in a visit, the trained model $model$. The threshold $\delta$, and $\gamma$.
                   \ENSURE The label of the subject.
                   \STATE $data_{input}$=$data_h$ concatenates $data_{base}$
                   \WHILE {True}
                   \STATE $result_{pred}$, $a_{pred}$=$model.predict(data_{input})$
                   \FOR{$ i=0  \quad to  \quad len( result_{pred}) $}
                   \IF{$result_{pred}[i]>=\delta[i]$}
                    \RETURN i    \qquad  // When $i==len(result_{pred})-1$, the result is representing unknown
                   \ENDIF
                   \ENDFOR
                   \STATE $is\_concat\_new\_data=False$
                   \FOR{$ i=0  \quad to  \quad len(a_{pred}) $}
                  \IF{$a_{pred}[i]>=\gamma[i]$}
                  \IF{The $ith$ examination is able to execute by medical institution}
                  \STATE $data_{input}$=$data_{input}$ concat $data_{ith}$
                  \STATE $is\_concat\_new\_data=True$
                  %\STATE
                  \ENDIF
                  \ENDIF
                \ENDFOR
                \IF{not $is\_concat\_new\_data$}
                \STATE Select a less cost and common examination $jth$ examination which does not execute in this visit and is able to execute by the medical institution.
                \IF{$jth$ examination is selected}
                \STATE $data_{input}$=$data_{input}$ concat $data_{jth}$
                \STATE $is\_concat\_new\_data=True$
                \ENDIF
                \ENDIF
                \IF{not $is\_concat\_new\_data$}
                 \RETURN unknown
                \ENDIF
                   \ENDWHILE
                   %\STATE Return Unknown
         \end{algorithmic}
\end{algorithm}

\section{Results}
\label{sec:Results}

\subsection{Human subjects}
\label{Human subjects}
Data used in the preparation of this article were obtained from the Alzheimer's Disease Neuroimaging Initiative (ADNI) dataset (\url{http://adni.loni.usc.edu}). We included $2,127$ subjects from ADNI $1$, ADNI GO, ADNI $2$, and ADNI $3$ for $9,593$ visits based on data availability. Detailed information can be found in Section S1. The characteristics of the subjects are shown in Table~\ref{ptd}. All the models will be verified in both a closed clinical setting and a real-world clinical setting. The configurations of the closed clinical setting and real-world clinical setting are as follows:

\begin{table*}[ht]
\centering
\caption{\textbf{Characteristics of subjects.} (Please note that visit data for the same subject should not be included in the training set, validation set, and test set simultaneously. However, data from some subjects in different time periods may belong to different disease categories, which can be considered as if they come from different subjects and are allowed to appear in separate datasets. As a result, the combined number of subjects in AD (Alzheimer's disease), CN (Cognitively Normal), MCI (Mild Cognitive Impairment), and SMC (Subjective Memory Concern) exceeds the actual number of subjects (in fact, there is subject overlap))\label{ptd}}
\begin{tabular}{cccccc}
\hline
                                                                           &                   & Data set & Training set & Validation set & Test set \\ \hline
\multirow{6}{*}{Age}                                                       & 54-59.9           & 80       & 36           & 2              & 59       \\
                                                                           & 60-69.9           & 596      & 246          & 10             & 442      \\
                                                                           & 70-70.9           & 1048     & 528          & 46             & 695      \\
                                                                           & 80-80.9           & 395      & 213          & 14             & 259      \\
                                                                           & 90-91.9           & 6        & 1            & 1              & 4        \\
                                                                           & Missing           & 2     & 1          & 0             & 1      \\ \hline
\multirow{2}{*}{Gender}                                                    & Female            & 1130     & 560          & 44             & 785      \\
                                                                           & Male              & 997      & 465          & 29             & 675      \\ \hline
\multirow{5}{*}{Educate}                                                                    & 8-10              & 40       & 18           & 2              & 23       \\
                                                                           & 11-13             & 353      & 176          & 13             & 243      \\
                                                                           & 14-16             & 823      & 403          & 26             & 558      \\
                                                                           & 17-20             & 900      & 424          & 32             & 628      \\
                                                                           & Missing           & 11     & 4          & 0             & 8      \\ \hline

\multirow{3}{*}{Ethnic category}                                                     & Hisp/Latino       & 73       & 32           & 5              & 49       \\
                                                            & Not Hisp/Latino   & 2042     & 986          & 67             & 1404     \\
                                                            & Missing           & 12       & 7            & 1              & 7        \\ \hline               
\multirow{7}{*}{\begin{tabular}[c]{@{}c@{}}Racial\\ category\end{tabular}} & Asian             & 40       & 20           & 0              & 25       \\
                                                                           & Black             & 88       & 41           & 5              & 57       \\
                                                                           & Hawaiian/Other PI & 2        & 0            & 0              & 2        \\
                                                                           & More than one     & 25       & 10           & 0              & 18       \\
                                                                           & White             & 1964     & 954          & 68             & 1350     \\
                                                                           & Am Indian/Alaskan & 4        & 0            & 0              & 4        \\
                                                                           & Missing           & 4        & 0            & 0              & 4        \\ \hline
\multirow{5}{*}{Marriage}                                                  & Married           & 1618     & 805          & 59             & 1100     \\
                                                                           & Never\_married    & 73       & 30           & 3              & 48       \\
                                                                           & Widowed           & 238      & 114          & 8              & 165      \\
                                                                           & Divorced          & 191      & 75           & 3              & 141      \\
                                                                           & Missing           & 7        & 1            & 0              & 6        \\ \hline
\multirow{4}{*}{Category}                                                  & AD                & 740      & 587          & 44             & 109      \\
                                                                           & CN                & 589      & 466          & 31             & 92       \\
                                                                           & MCI               & 1082     & 0            & 0              & 1082     \\
                                                                           & SMC               & 280      & 0            & 0              & 280      \\ \hline 
\end{tabular}
\centering

\end{table*}

\subsubsection{Closed world setting}
\label{Closed world setting}

In the closed world setting, $85\%$ of AD and CN subjects were used for the training set, $5\%$ of AD and CN subjects were used for the validation set, and $20\%$ of AD and CN subjects were used for the test set.
% ===========================================================
%Thus, for a subject, $443,795$ strategies were generated.
\subsubsection{Real world setting}
\label{Real world setting}
A total of  $2127$ subjects with $9,593$ visits were included in our work. A subject during a visit may require different categories of examination. Every combination of those examinations represents a diagnostic strategy. Thus, for the data of subject, 443,795 strategies were generated. These AD and CN subjects were randomly assigned to the training, validation, and test sets. The training set contained $1,025$ subjects with $3,986$ visits and generated $180,682$ strategies. In the training set, $587$ subjects with $1,781$ visits were AD and developed $80,022$ strategies, and $466$ subjects with $2,205$ visits were CN and generated $100,660$ strategies. The validation set contained $73$ subjects with $254$ visits and generated $11,898$ strategies. In the validation set, $44$ subjects with $127$ visits were AD and develop $6,008$ strategies, and $31$ subjects with $127$ visits were CN and generated $5,890$ strategies. The test set contained $1,460$ subjects with $5,353$ visits. In the test set, $109$ subjects with $305$ visits were AD, $92$ subjects with $411$ visits were CN, $1,082$ subjects with $4,357$ visits were MCI (mild cognitive impairment), and $280$ subjects with $280$ visits were SMC (significant memory concern). Notably, the dataset contains multiple visits for a subject's progression from CN to AD.\par
We note that the lack of examination data was able to simulate the lack of the executive ability of the examination by the medical institution. It is logically equivalent to missed examinations in medical institutions. All subjects with labels containing at least one of the above categories of information were considered in this study.

\subsection{Comparison methods}
We validated the effectiveness of OpenclinicalAI by comparing its performance  with recent work: image-based models (DSA-3D-CNN (2016)~\cite{hosseini2016alzheimer}, VoxCNN-ResNet (2017)~\cite{korolev2017residual}, CNN-LRP (2019)~\cite{bohle2019layer}, Dynamic-image-VGG (2020)~\cite{xing2020dynamic}, Ncommon-MRI (2022)~\cite{qiu2022multimodal}, DenseNet-XGBoost) and multimodal input models (FCN-MLP (2020)~\cite{qiu2020development}). In addition, the transfer learning-based model DenseNet-XGBoost was the previous state-of-the-art baseline model, since among the recent AI diagnosis studies, the transfer learning framework of the pretrained model achieved state-of-the-art performance in many diagnosis tasks based on medical images~\cite{lee2019explainable,esteva2017dermatologist,tschandl2020human,poplin2018prediction,kermany2018identifying}.
% LEAR~\cite{oh2022learn}
\subsection{Experimental setup}
The model was optimized using mini-batch stochastic gradient descent with Adam and a base learning rate of $0.0005$~\cite{kingma2015adam}. All comparison models were constructed and trained according to the needs of AD diagnosis tasks and their official codes under the same settings. The experiments were conducted on a Linux server equipped with Tesla P$40$ and Tesla P$100$ GPUs.

\begin{table*}[ht]
\centering
\small 
\caption{\textbf{Model performance in closed clinical setting setting.}\label{close_model_performance}}
\resizebox{\linewidth}{!}{
\begin{tabular}{lcccc}
\hline
Model &  AUC(95\% CI) & Accuracy(95\% CI) & Sensitivity(95\% CI) & Specificity(95\% CI) \\
\hline
    DSA-3D-CNN~\cite{hosseini2016alzheimer} & 89.23\%(87.91\%-90.56\%)  & 84.20\%(82.76\%-85.60\%)   & 77.05\%(74.55\%-79.49\%)  & 89.00\%(87.46\%-90.66\%)\\
    VoxCNN-ResNet~\cite{korolev2017residual} & 90.64\%(89.31\%-91.80\%)  & 85.24\%(83.76\%-86.56\%)   & 80.76\%(78.44\%-83.04\%)  & 88.65\%(86.96\%-90.25\%)\\
    CNN-LRP~\cite{bohle2019layer} & 93.21\%(92.11\%-94.16\%)  & 82.36\%(80.72\%-83.76\%)   & 90.68\%(88.91\%-92.27\%)  & 75.97\%(73.57\%-78.21\%)\\
    Dynamic-image-VGG~\cite{xing2020dynamic} & 83.22\%(81.58\%-84.81\%)  & 73.42\%(71.71\%-75.12\%)   & 85.24\%(83.11\%-87.34\%)  & 64.41\%(61.98\%-66.89\%)\\
    % LEAR~\cite{oh2022learn} & 000  & 000   & 000  & 000\\
    Ncomms2022-MRI~\cite{qiu2022multimodal} & 93.06\%(92.04\%-94.05\%)  & 87.40\%(86.12\%-88.68\%)   & 82.53\%(80.26\%-84.71\%)  & 91.08\%(89.66\%-92.57\%)\\
    DenseNet-XGBoost & 97.79\%(97.30\%-98.27\%)  & 93.12\%(92.12\%-94.08\%)   & 91.09\%(89.40\%-92.90\%)  & 94.53\%(93.35\%-95.64\%)\\
    FCN-MLP~\cite{qiu2020development} & 97.28\%(96.71\%-97.81\%)  & 90.72\%(89.56\%-91.84\%)   & \textbf{92.26\%(90.59\%-93.76\%)}  & 89.55\%(87.89\%-91.10\%)\\
    \textbf{OpenClinicalAI} & \textbf{99.50\%(99.11\%-99.80\%)}  & \textbf{96.52\%(95.76\%-97.20\%)}   & 91.86\%(90.13\%-93.37\%)  & \textbf{100\%(100\%-100\%)}\\
\hline
\end{tabular}
}
\begin{tablenotes}
        \footnotesize
        \item[1] CI = confidence interval. To evaluate the evaluation index of the AI model, a non-parametric bootstrap method was applied to calculate the CI  for the evaluation index~\cite{efron1994introduction}. We calculated $95\%$ CI for every evaluation index. We randomly sampled $2,500$ cases from the test set and evaluated the AI model by the sampled set for every evaluation index. $2,000$ repeated trials were executed, and $2,000$ values of the evaluation index were generated. The $95\%$ CI was obtained by the $2.5$ and $97.5$ percentiles of the distribution of the evaluation index values.
  \end{tablenotes}
\end{table*}

%=============thr=0.5-0.75
\begin{table*}[ht]
\centering
\caption{Model performance in real-world clinical setting.\label{real_model_performance}}
\resizebox{\linewidth}{!}{
\begin{tabular}{p{4.5cm}ccccc}
\hline
  \multirow{2}{*}{Model}&
  \multicolumn{2}{c}{AUC(95\% CI)}&\multicolumn{3}{c}{Sensitivity(95\% CI)}\cr
  &AD & CN &AD & CN & unknown\cr
    \hline
    DSA-3D-CNN-Thr & 78.90\%(74.91\%-82.71\%)  & 66.72\%(63.33\%-70.07\%)  & 75.93\%(68.66\%-82.92\%)& 85.56\%(80.12\%-90.50\%)  & 6.49\%(5.51\%-7.63\%)\\
    DSA-3D-CNN-OpenMax & 78.03\%(74.03\%-81.62\%)  & 66.86\%(63.41\%-70.38\%)   & 60.29\%(51.97\%-68.38\%)& 72.63\%(65.86\%-79.25\%)  & 30.26\%(28.46\%-32.22\%)\\
    VoxCNN-ResNet-Thr & 76.02\%(71.66\%-80.07\%)  & 66.42\%(63.01\%-69.77\%)   & 72.82\%(65.41\%-80.00\%) & 85.20\%(80.09\%-90.15\%)  & 7.05\%(6.00\%-8.12\%)\\
    VoxCNN-ResNet-OpenMax & 76.33\%(72.32\%-79.97\%)  & 64.12\%(60.58\%-67.60\%)   & 59.86\%(51.67\%-67.39\%) & 65.02\%(58.06\%-72.13\%)  & 29.14\%(27.33\%-31.05\%)\\
    CNN-LRP-Thr & 82.93\%(79.43\%-86.25\%)  & 67.92\%(64.76\%-71.14\%)   & 89.51\%(83.82\%-94.24\%) & 71.59\%(64.67\%-78.02\%)  & 6.36\%(5.34\%-7.45\%)\\
    CNN-LRP-OpenMax & 82.82\%(78.96\%-86.18\%)  & 67.91\%(64.72\%-70.91\%)   & 85.29\%(78.73\%-90.90\%) & 58.13\%(50.80\%-65.13\%)  & 20.16\%(15.11\%-23.89\%)\\
    Dynamic-image-VGG-Thr & 71.75\%(67.55\%-75.64\%)  & 63.07\%(59.66\%-66.53\%)   & 85.25\%(79.35\%-90.72\%)& 64.16\%(57.29\%-70.93\%)  & 0.04\%(0.00\%-0.13\%)\\
    Dynamic-image-OpenMax & 70.75\%(66.62\%-74.84\%)  & 63.28\%(59.76\%-66.63\%)   & 79.19\%(72.35\%-85.82\%)& 57.28\%(50.26\%-64.40\%)  & 14.26\%(12.78\%-15.77\%)\\
    % LEAR~\cite{oh2022learn} & 000  & 000   & 000 & 000  & 000\\
    Ncomms2022-MRI-Thr  & 81.15\%(77.64\%-84.44\%)  & 69.32\%(65.94\%-72.40\%)   & 79.13\%(72.43\%-85.48\%)& 87.84\%(82.98\%-92.59\%)  & 10.34\%(9.14\%-11.64\%)\\
    Ncomms2022-MRI-OpenMax  & 81.21\%(77.55\%-84.74\%)  & 69.33\%(66.11\%-72.46\%)   & 63.98\%(56.12\%-71.87\%)& 68.57\%(62.05\%-75.28\%)  & 38.14\%(36.13\%-40.25\%)\\
    DenseNet-XGBoost-Thr & 84.00\%(80.55\%-87.28\%)  & 87.79\%(85.06\%-90.25\%)  &54.83\%(46.04\%-63.01\%) & 33.33\%(26.63\%-39.79\%)  &  88.88\%(87.53\%-90.18\%)\\
    FCN-MLP-Thr & 91.71\%(89.42\%-93.78\%)  & 74.16\%(71.11\%-76.98\%)   & 91.93\%(86.79\%-96.15\%)& \textbf{90.00\%(85.11\%-94.08\%)}  & 14.64\%(13.22\%-16.09\%)\\
    FCN-MLP-OpenMax & 92.55\%(90.26\%-94.53\%)  & 74.30\%(71.45\%-77.04\%)   & 73.15\%(65.67\%-80.45\%)& 66.47\%(59.45\%-73.17\%)  & 57.95\%(55.81\%-60.03\%)\\
    OpenClinicalAI-G & 85.54\%(83.30\%-87.70\%)  &  82.65\%(80.05\%-85.30\%)   & \textbf{93.28\%(88.66\%-96.92\%)} & 76.84\%(70.79\%-82.75\%)  & 34.33\%(32.21\%-36.30\%)\\
    \textbf{OpenClinicalAI} &\textbf{95.02\%(93.04\%-96.62\%)}  &\textbf{99.27\%(98.54\%-99.81\%)}   &84.92\%(78.91\%-90.51\%) &81.27\%(75.51\%-86.67\%)  &\textbf{93.96\%(92.90\%-94.92\%)}\\
\hline
\end{tabular}
}

\end{table*}

\begin{table*}[ht]
\centering
\caption{The number of examination for diagnosis in test set.\label{number_of_examination}}
\resizebox{\linewidth}{!}{
\begin{tabular}{cp{4.5cm}ccccccccccccc}
\multicolumn{2}{c}{Model}                                                       & Base      & Cog       & CE   & Neur & FB   & PE   & Blood & Urine & MRI       & FDG & AV45 & Gene & CSF \\\hline
\multirow{8}{*}{\begin{tabular}[c]{@{}c@{}}closed \\  world\\   setting\end{tabular}} & DSA-3D-CNN~\cite{hosseini2016alzheimer}                               & 0         & 0         & 0    & 0    & 0    & 0    & 0     & 0     & 594 (122) & 0   & 0    & 0    & 0   \\
                                     & VoxCNN-ResNet~\cite{korolev2017residual}                            & 0         & 0         & 0    & 0    & 0    & 0    & 0     & 0     & 594 (122) & 0   & 0    & 0    & 0   \\
                                     & CNN-LRP~\cite{bohle2019layer}                                  & 0         & 0         & 0    & 0    & 0    & 0    & 0     & 0     & 594 (122) & 0   & 0    & 0    & 0   \\
                                     & Dynamic-image-VGG~\cite{xing2020dynamic}                        & 0         & 0         & 0    & 0    & 0    & 0    & 0     & 0     & 594 (122) & 0   & 0    & 0    & 0   \\
                                    %  & LEAR~\cite{oh2022learn}                                     & 0         & 0         & 0    & 0    & 0    & 0    & 0     & 0     & 594 (122) & 0   & 0    & 0    & 0   \\
                                    &Ncomms2022-MRI~\cite{qiu2022multimodal} & 0         & 0         & 0    & 0    & 0    & 0    & 0     & 0     & 594 (122) & 0   & 0    & 0    & 0   \\
                                     & DenseNet-XGBoost                             & 0         & 0         & 0    & 0    & 0    & 0    & 0     & 0     & 594(122)       & 0   & 0    & 0    & 0   \\
                                     & FCN-MLP~\cite{qiu2020development}                                  & 716 & 716 & 0    & 0    & 0    & 0    & 0     & 0     & 594 (122) & 0   & 0    & 0    & 0   \\
                                     & \textbf{OpenClinicalAI}                           & 716       & 216       & 145  & 114  & 144  & 137  & 34    & 32    & 71        & 28  & 28   & 1    & 0   \\\hline
\multirow{15}{*}{\begin{tabular}[c]{@{}c@{}}real\\  world\\   setting\end{tabular}} & DSA-3D-CNN-Thr                               & 0         & 0         & 0    & 0    & 0    & 0    & 0     & 0     & 4609(744)      & 0   & 0    & 0    & 0   \\
& DSA-3D-CNN-OpenMax                               & 0         & 0         & 0    & 0    & 0    & 0    & 0     & 0     & 4609(744)      & 0   & 0    & 0    & 0   \\
                                     & VoxCNN-ResNet-Thr                            & 0         & 0         & 0    & 0    & 0    & 0    & 0     & 0     & 4609(744)      & 0   & 0    & 0    & 0   \\
                                     & VoxCNN-ResNet-OpenMax                            & 0         & 0         & 0    & 0    & 0    & 0    & 0     & 0     & 4609(744)      & 0   & 0    & 0    & 0   \\
                                     & CNN-LRP-Thr                                  & 0         & 0         & 0    & 0    & 0    & 0    & 0     & 0     & 4609(744)      & 0   & 0    & 0    & 0   \\
                                     & CNN-LRP-OpenMax                          & 0         & 0         & 0    & 0    & 0    & 0    & 0     & 0     & 4609(744)      & 0   & 0    & 0    & 0   \\
                                     & Dynamic-image-VGG-Thr                        & 0         & 0         & 0    & 0    & 0    & 0    & 0     & 0     & 4609(744)      & 0   & 0    & 0    & 0   \\
                                     & Dynamic-image-VGG-OpenMax                        & 0         & 0         & 0    & 0    & 0    & 0    & 0     & 0     & 4609(744)      & 0   & 0    & 0    & 0   \\
                                    %  & LEAR~\cite{oh2022learn}                                     & 0         & 0         & 0    & 0    & 0    & 0    & 0     & 0     & 4609(744)      & 0   & 0    & 0    & 0   \\
                                    &Ncomms2022-MRI-Thr & 0         & 0         & 0    & 0    & 0    & 0    & 0     & 0     & 4609(744)      & 0   & 0    & 0    & 0   \\
                                    &Ncomms2022-MRI-OpenMax & 0         & 0         & 0    & 0    & 0    & 0    & 0     & 0     & 4609(744)      & 0   & 0    & 0    & 0   \\
                                     & DenseNet-XGBoost                             & 0         & 0         & 0    & 0    & 0    & 0    & 0     & 0     & 4609(744)      & 0   & 0    & 0    & 0   \\
                                     & FCN-MLP-Thr                                  & 5353      & 5353      & 0    & 0    & 0    & 0    & 0     & 0     & 4609(744)      & 0   & 0    & 0    & 0   \\
                                     & FCN-MLP-OpenMax                          & 5353      & 5353      & 0    & 0    & 0    & 0    & 0     & 0     & 4609(744)      & 0   & 0    & 0    & 0   \\
                                     & OpenClinicalAI-G & 5353      & 2921      & 2483 & 2261 & 2388 & 2245 & 510   & 228   & 1533      & 684 & 509  & 347  & 232 \\
                                     & \textbf{OpenClinicalAI}                           & 5353      & 5353      & 2146 & 2051 & 2021 & 1986 & 450   & 154   & 1663      & 681 & 444  & 449  & 240\\\hline
\end{tabular}
}
 \begin{tablenotes}
         \footnotesize
         \item The $594(122)$ in the column of MRI means that of all the samples required to provide MRI, only $594$ samples provided information, and 122 samples did not.
   \end{tablenotes}
\end{table*}

\subsection{The performance  in the closed clinical setting}

\begin{figure*}
\centering
\begin{minipage}[a]{0.49\textwidth}
%\caption*{a.}
\leftline{\textbf{a}}
\includegraphics[height=7.5cm, width=9cm]{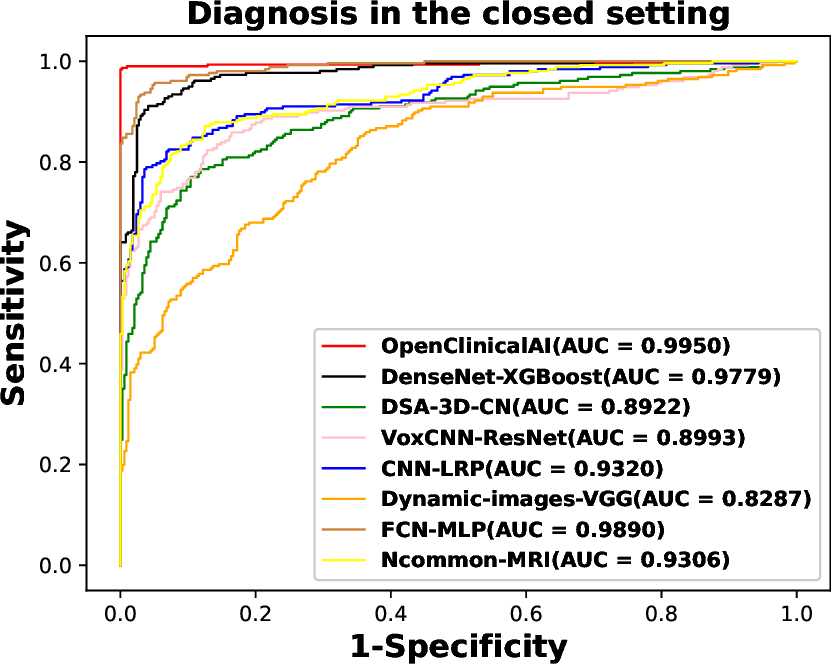} 
\end{minipage}
%\hspace{0.5cm}
\begin{minipage}[a]{0.49\textwidth}
\leftline{\textbf{b}}
\includegraphics[height=7.5cm, width=9cm]{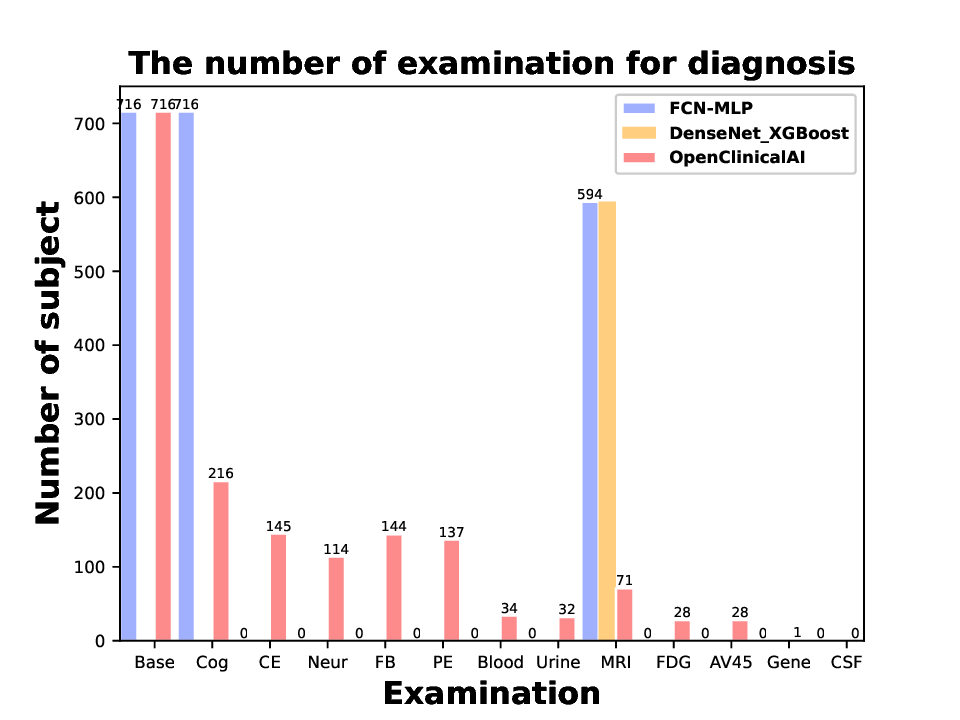} 
\end{minipage}\\
\quad
\caption{\textbf{The workflow of the baseline clinical AI system and OpenClinicalAI. a,} The performance of OpenClinicalAI against recent works (DSA-3D-CNN, VoxCNN-ResNet, CNN-LRP, Dynamic-image-VGG, FCN-MLP, Ncommon-MRI, DenseNet-XGBoost) in a closed clinical setting. \textbf{b,} The number of clinical examinations used by OpenClinicalAI on the test set against the state-of-the-art model (such as FCN-MLP, DenseNet-XGBoost).
\label{performance_closed}}
\end{figure*}

As shown in Table~\ref{close_model_performance} and Fig.~\ref{performance_closed} (a), in general, the performance of multimodal model (such as AUC score of FCN-MLP achieving $97.28\%$ [$95\%$ CI $96.71\%$-$97.81\%$], accuracy score achieving $90.72\%$ [$95\%$ CI $89.56\%$-$91.84\%$]) is better than that of single-modal model (such as AUC score of CNN-LRP achieving $93.21\%$ [$95\%$ CI $92.11\%$-$94.16\%$], accuracy score achieving $82.36\%$ [$95\%$ CI $80.72\%$-$83.76\%$]). The OpenClinicalAI achieves the state-of-the-art performance with the AUC score of $99.50\%$ [$95\%$ CI $99.11\%$-$99.80\%$] and the accuracy score of $96.52\%$ [$95\%$ CI $95.76\%$-$97.20\%$] in the closed setting. Considering the confidence interval of the model, OpenClincialAI does not show significant improvement compared with other multimodal-based models. This  indicates that the task of diagnosing AD in a closed clinical setting is not very challenging, and there is not much room for improvement~\cite{tanveer2020machine,mahajan2020machine}.

The essential improvement from the state-of-the-art model to OpenClinicalAI is that the latter can dynamically develop personalized diagnosis strategies according to specific subjects and medical institutions. As shown in Table~\ref{number_of_examination} and Fig.~\ref{performance_closed} (b), less than $10\%$ of the subjects require a nuclear magnetic resonance scan, and most of the subjects only require less demanding examination, such as cognitive examination. We conclude that OpenClinicalAI can avoid unnecessary examinations for subjects and is suitable for medical institutions with varying examination facilities \footnote{Different hospitals have various clinical settings, such as community hospitals without nuclear magnetic resonance machines and large hospitals with multiple facilities}. Of the $716$ samples in the test set, only $594$ samples had MRI data, and the remaining $122$ samples were simply discarded because they could not be used as inputs to the corresponding MRI model.

\subsection{The performance of AD diagnosis in the real-world clinical setting}

To apply the model to the real-world clinical setting, the model trained in a closed clinical setting usually needs to set a threshold or add OpenMax or use the generated pattern to identify samples of unknown categories~\cite{geng2020recent}. As shown in Table~\ref{real_model_performance} and Fig.~\ref{performance_open}, in general, the performance of all models has declined significantly (AUC score decline for AD was in the range of [$4.7\%$-$14.6\%$], accompanied by a very low sensitivity to unknown), but the multimodal model is still better than the single-modal model. OpenClinicalAI achieves state-of-the-art performance (AUC score to AD reaching $95.02\%$ [$95\%$ CI $93.04\%$-$96.62\%$], sensitivity to unknown reaching $93.96\%$ [$95\%$ CI $92.90\%$-$94.92\%$]). In addition, except for our models, all models have high recognition accuracy for samples of known categories and significantly low recognition accuracy for samples of unknown categories (for example, the sensitivity of FCN-MLP-Thr to AD is $91.93\%$ [$95\%$ CI $86.79\%$-$96.15\%$], the sensitivity to CN is $90.00\%$ [$95\%$ CI $85.11\%$-$94.08\%$], and the sensitivity to unknown was $14.64\%$ [$95\%$ CI $13.22\%$-$16.09\%$]), or vice versa (the sensitivity of DenseNet-XGBoost-Thr to unknown is $88.88\%$ [$95\%$ CI $87.53\%$-$90.18\%$]). 
% and Fig. ~\ref{performance_open2}

\newcommand{\mysize}{4.5cm}

\begin{figure*}[h]%[htbp]
\centering
\subfigure[]{
\includegraphics[width=\mysize]{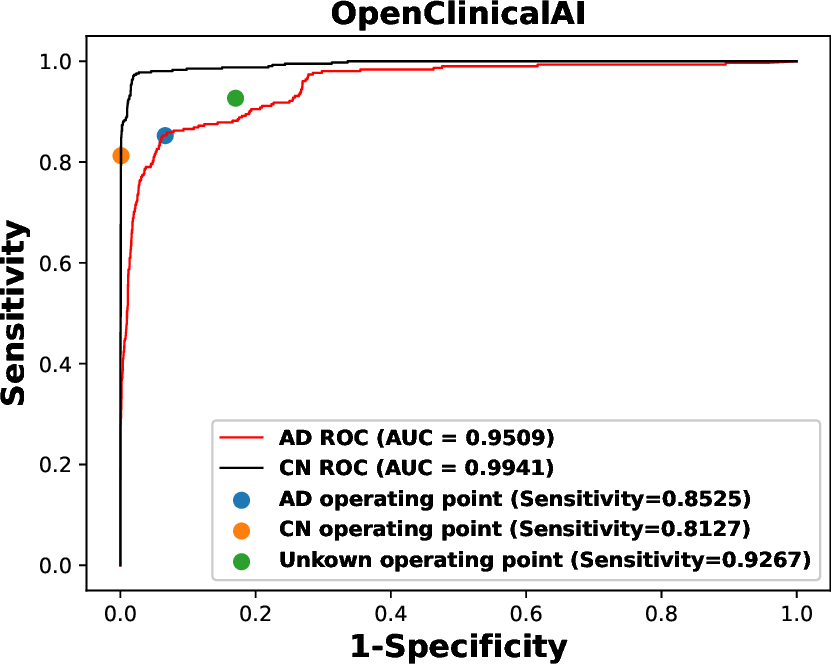}
}
\quad
\subfigure[]{
\includegraphics[width=\mysize]{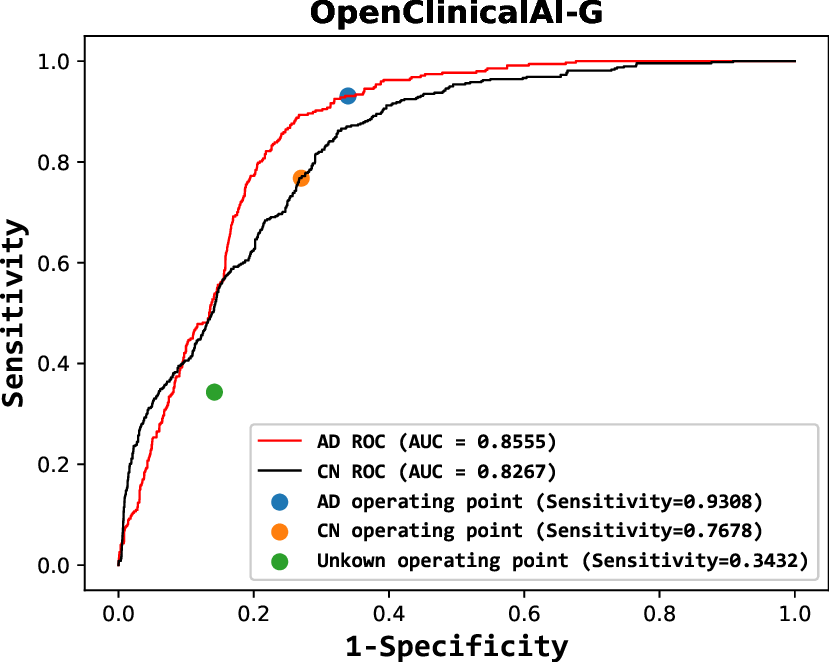}
}
% \quad
% \subfigure[]{
% \includegraphics[width=\mysize]{image/dynamic_AI.eps}
% }
\quad
\subfigure[]{
\includegraphics[width=\mysize]{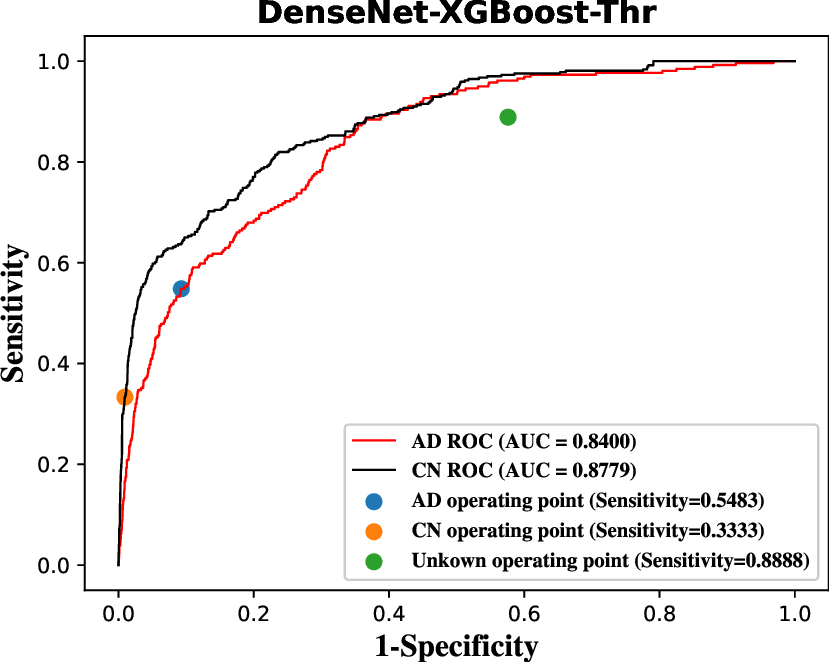}
}

\quad
\subfigure[]{
\includegraphics[width=\mysize]{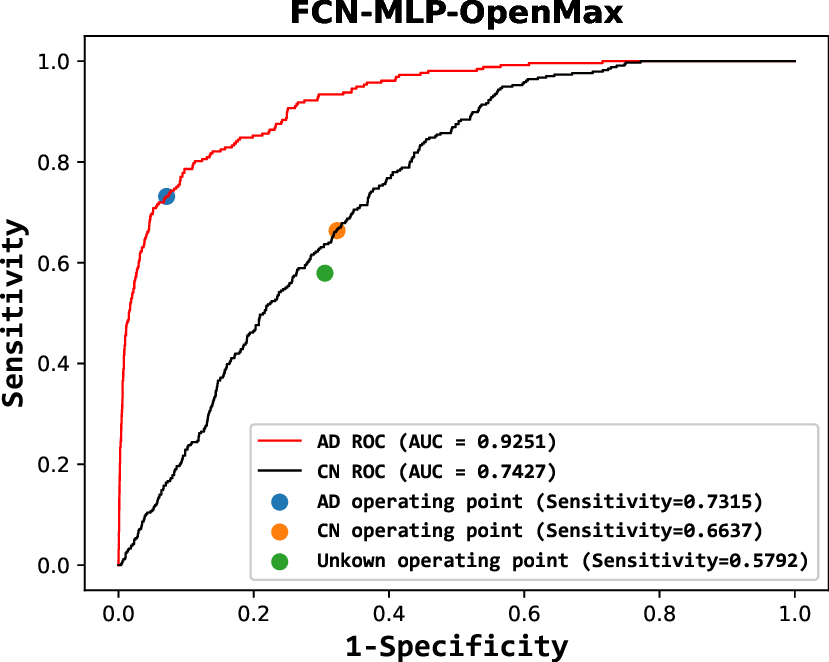}
}
\quad
\subfigure[]{
\includegraphics[width=\mysize]{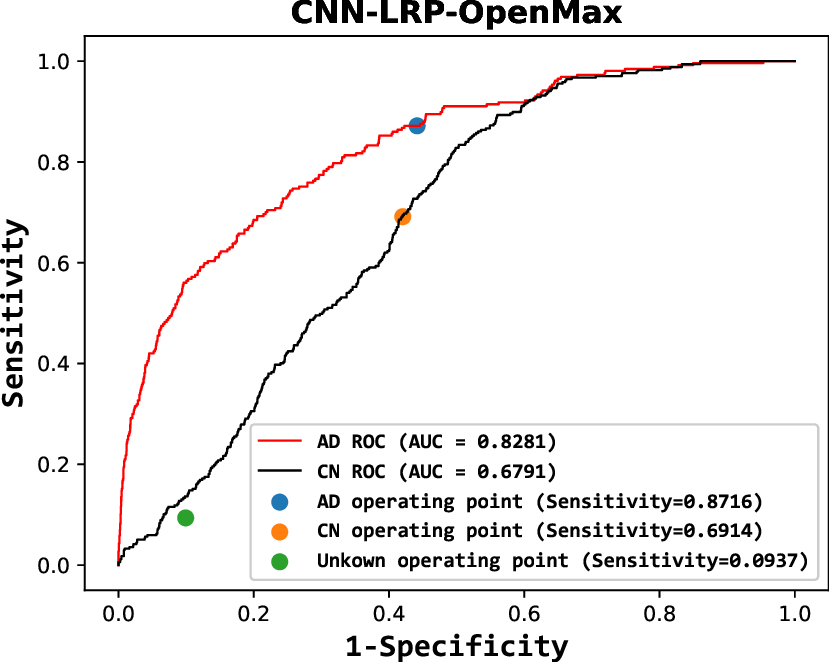}
}
\quad
\subfigure[]{
\includegraphics[width=\mysize]{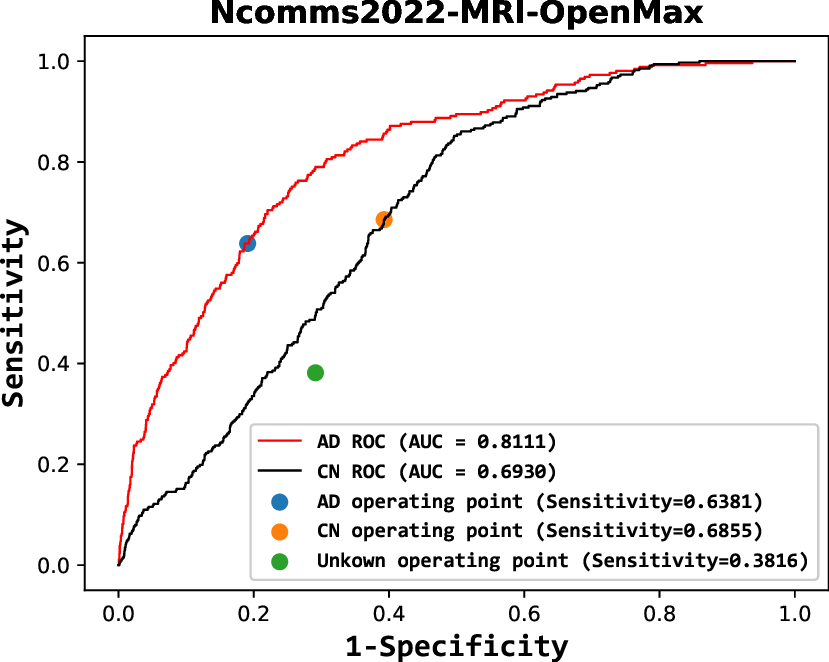}
}
\quad
\subfigure[]{
\includegraphics[width=\mysize]{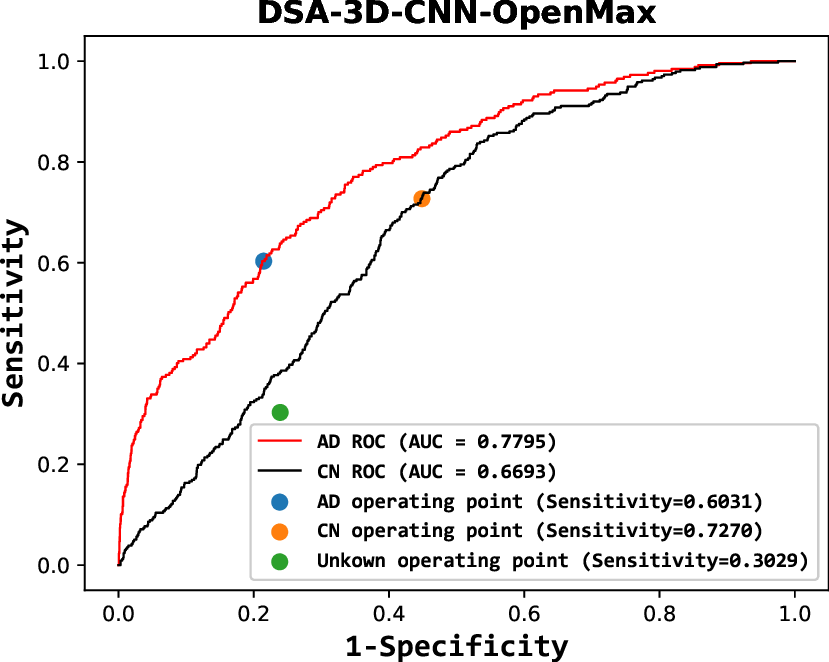}
}
\quad
\subfigure[]{
\includegraphics[width=\mysize]{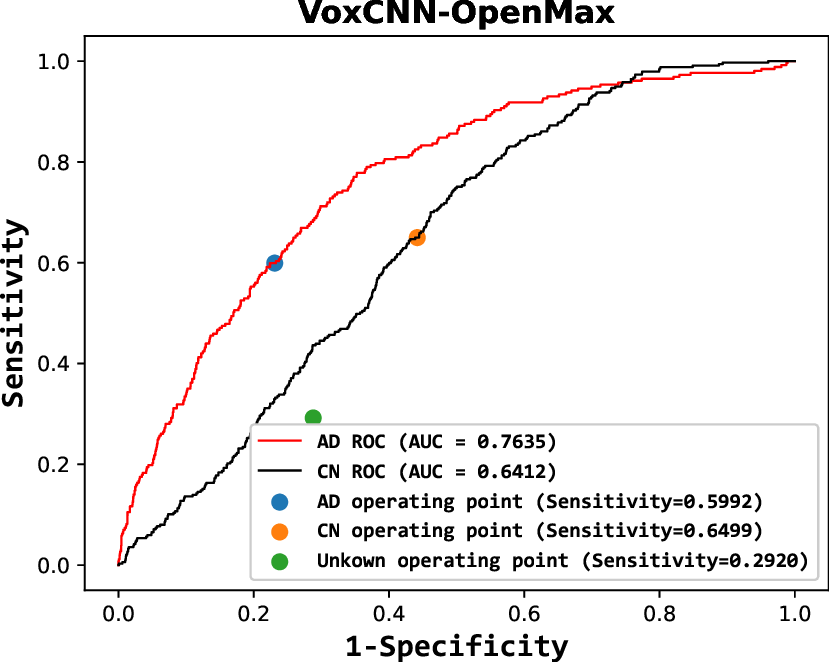}
}
\quad
\subfigure[]{
\includegraphics[width=5cm]{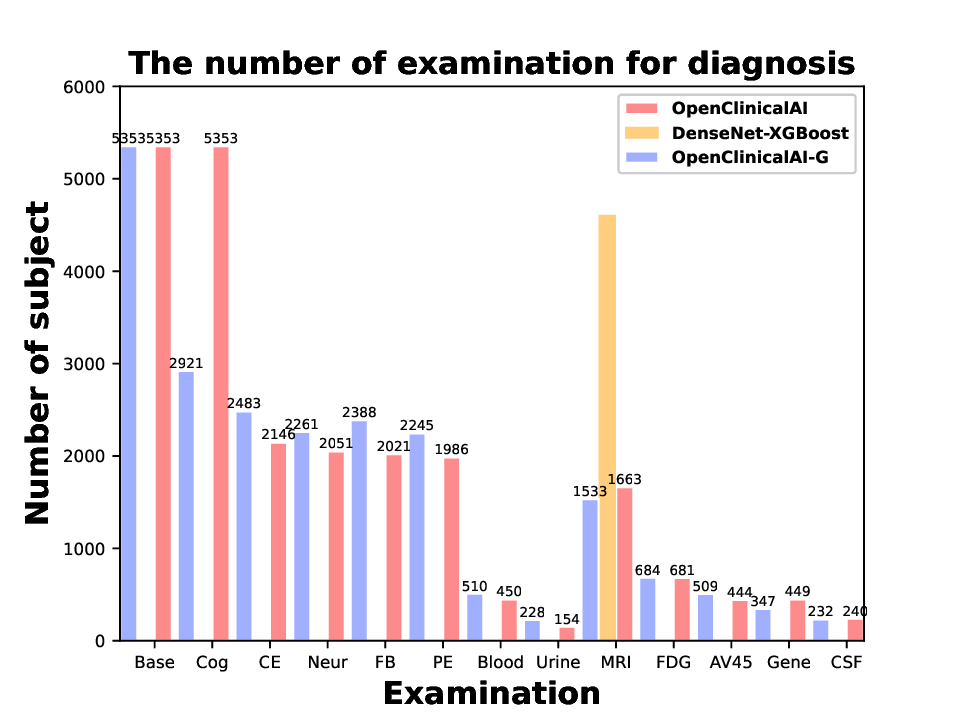}
}
% \quad
% \subfigure[]{
% \includegraphics[width=\mysize]{image/open_set_methond.eps}
% }
\caption{\textbf{The performance of OpenClinicalAI with personalized strategies against the state-of-the-art model in the real-world setting. }  \label{performance_open}}
\end{figure*}

% As shown in Figure. S~\ref{performance_open} a, b, and c
Compared to the state-of-the-art model, OpenClinicalAI demonstrates a significant improvement in the AUC of identification of AD subjects $(+2.47\%)$ and the AUC of identification of CN subjects $(+11.48\%)$. It is worth noting that, as shown in Fig.~\ref{performance_open} (a), OpenClinicalAI has a substantial improvement in the sensitivity of AD, CN, and unknown operating points (the sensitivity of OpenClinicalAI to unknown is $93.96\%$ [$95\%$ CI $92.90\%$-$94.92\%$]).

For a state-of-the-art model, such as DenseNet-XGBoost, the sensitivity of known (AD and CN) subjects is low, the sensitivity of AD is just $54.83\%$, and the sensitivity of CN is just $33.33\%$. This indicates that most known subjects will be marked as unknown and sent to a clinician for diagnosis. Moreover, the sensitivity of unknown subjects is $88.88\%$, meaning $11.12\%$ of unknown subjects will be misdiagnosed. In addition, the DenseNet-XGBoost model requires that every subject has a nuclear magnetic resonance scan, and hence every medical institution that deploys the baseline model must be equipped with a nuclear magnetic resonance apparatus.

For OpenClinicalAI-G, which does not have an OpenMax mechanism but instead uses a generative-discriminative mechanism, the sensitivity for known (AD and CN) subjects is as good as that of OpenClinicalAI with an OpenMax mechanism~\cite{perera2020generative} (the sensitivity of OpenClinicalAI-G to AD is $85.54\%$ [$95\%$ CI $83.30\%$-$87.70\%$], and the sensitivity to CN is $82.65\%$ [$95\%$ CI $80.05\%$-$85.30\%$]). In contrast, the sensitivity of unknown subjects is much worse than that of OpenClinicalAI with an OpenMax mechanism (the sensitivity of OpenClinicalAI-G to unknown subjects is $34.33\%$ [$95\%$ CI $32.21\%$-$36.30\%$]). This means that most unknown subjects will be misdiagnosed, which is unacceptable in real-world settings. The generative model of open set recognition is not very helpful for AD diagnosis in a real-world clinical setting. In addition, as shown in Table~\ref{real_model_performance}, the combination of the OpenMax mechanism and traditional model cannot help AD diagnosis in a real-world clinical setting.

In contrast, OpenClinicalAI diagnoses most of the known (AD and CN) subjects correctly, marks most of the rest as unknown, and sends them to the clinician for further diagnosis. In addition, most unknown subjects are correctly identified, and the misdiagnosis of unknown subjects is only $6.04\%$. This means that OpenClinicalAI has significant potential application value for implementation in real-world settings. In addition, as shown in Fig.~\ref{performance_open} (i), similar to the behaviors of OpenClinicalAI in the closed setting, OpenClinicalAI can develop and adjust diagnosis strategies for every subject dynamically in the real-world setting. Only a small portion of subjects require a nuclear magnetic resonance scan and more costly (both in terms of economy and potential harm) examinations.

\subsection{Development of diagnostic strategies in the real-world clinical setting}

Unlike the current mainstream AD diagnostic models in which all subjects require a nuclear magnetic resonance scan, OpenClinicalAI develops personalized diagnostic strategies for each subject. For every subject, first, it will acquire the base information of the subject. Second, it will give a final diagnosis or receive other examination information according to the current data of the subject. Third, the previous step is repeated until the diagnosis is finalized or there is no further examination. 
%Table S~\ref{methond}
%the test set( Table S~\ref{institutions})
As shown in Fig.~\ref{strategy} (a), the diagnosis strategies of subjects are not the same (as shown in Supplementary Table S2). Our model dynamically develops $35$ diagnosis strategies according to different subject situations and all $40$ examination abilities of medical institutions in the test set (as shown in Supplementary Table S3). For the known (AD and CN) subjects, as shown in Fig.~\ref{strategy} (b) and (c), most of the subjects require low-cost examinations (such as cognition examination (CE)). A small portion of subjects require high-cost examinations (such as cerebral spinal fluid analysis (CSF)). For unknown subjects, as shown in Fig.~\ref{strategy} (d), different from the diagnosis of known (AD and CN) subjects, identifying unknown subjects is more complex and more dependent on high-cost examinations. The reason is that according to the mechanism of OpenClinicalAI, it will do its best to distinguish whether the subject belongs to the known categories. When it fails, it will mark the subject as unknown. This means that the unknown subject will undergo more examinations. The details of the high-cost examination requirements are as follows:
(1) $33.94\%$ of unknown subjects require a nuclear magnetic resonance scan (that of the known subject is $12.43\%$).
(2) $13.95\%$ of unknown subjects require a positron emission computed tomography scan with $18$-FDG (that of the known subject is $4.75\%$).
(3) $8.67\%$ of unknown subjects require a positron emission computed tomography scan with AV$45$ (that of the known subject is $5.87\%$).
(4) $9.38\%$ of unknown subjects require a gene analysis (that of the known subject is $1.96\%$).
(5) $5.13\%$ of unknown subjects require a cerebral spinal fluid analysis (that of the known subject is $0.28\%$).

\newcommand{\methodsize}{6cm}

\begin{figure*}[h]
\centering
\subfigure[]{
\includegraphics[width=\methodsize]{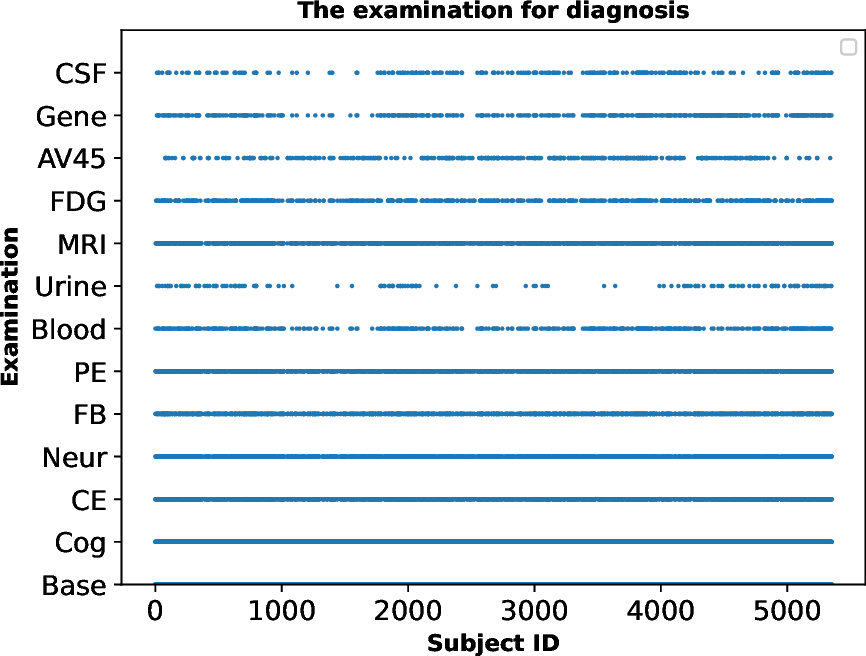}
}
\quad
\subfigure[]{
\includegraphics[width=\methodsize]{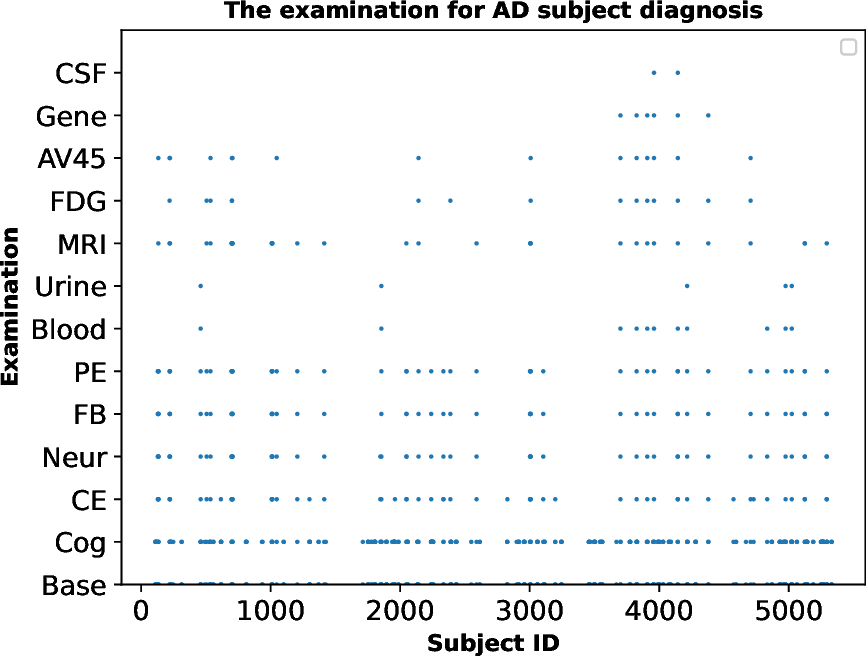}
%\caption{fig1}
}
\quad
\subfigure[]{
\includegraphics[width=\methodsize]{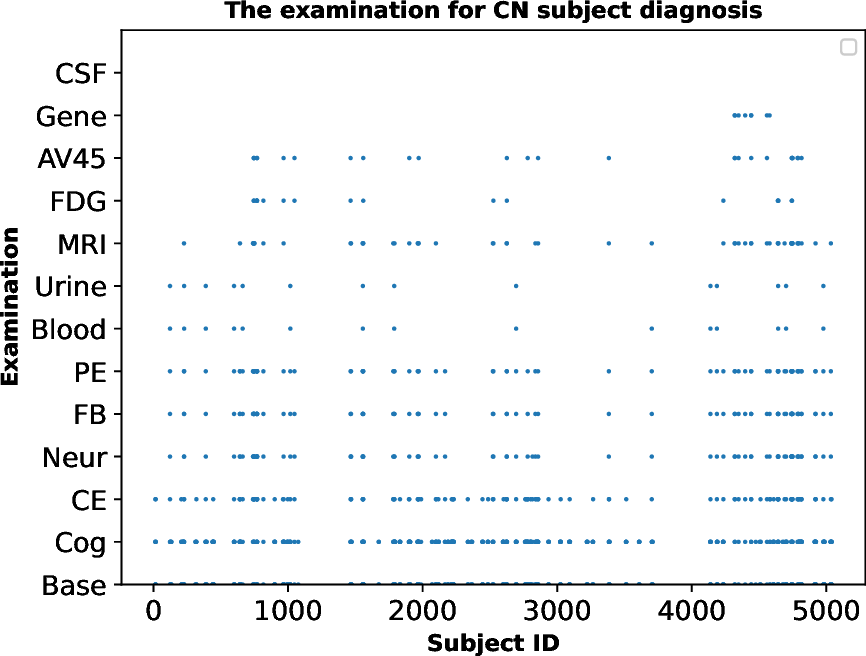}
}
\quad
\subfigure[]{
\includegraphics[width=\methodsize]{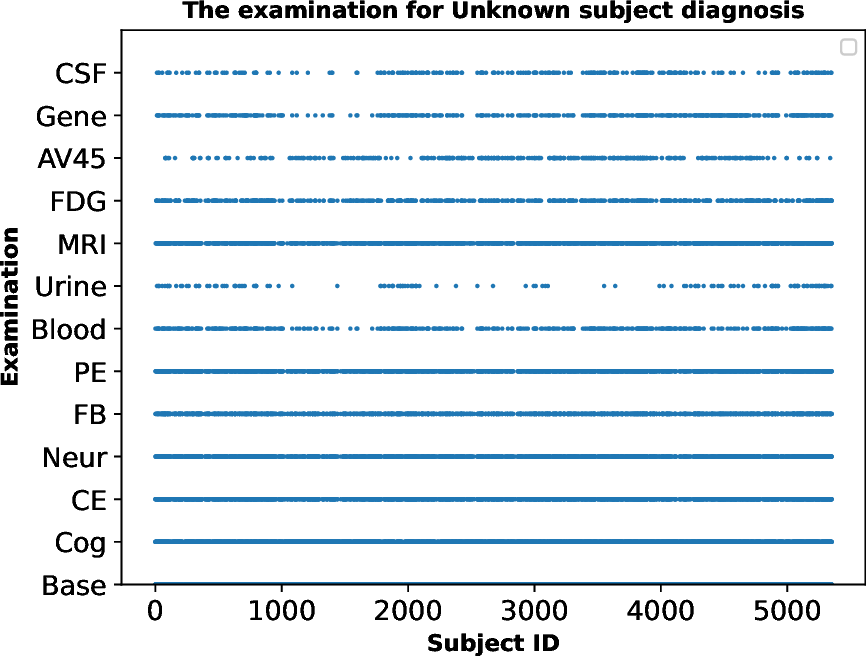}
}

  \caption{\textbf{Diagnosis strategies for subjects. a, } Diagnosis strategies for all subjects. Due to OpenClinicalAI developing and adjusting the examination for each subject, the selection of examinations for subjects is not the same. \textbf{b,} Diagnosis strategies for AD subjects. Compared to the high-cost examination, OpenClinicalAI pays more attention to the subject's basic information and cognitive, mental, behavioral, and physical examination information for the AD subject. In contrast, biochemical testing, imaging, and genetic data are less considered. \textbf{c,} Diagnosis strategies for CN subjects. The behaviors of OpenClinicalAI for CN recognition are similar to those for AD diagnosis, and the difference between those behaviors is that more examinations are required to identify the CN subject. \textbf{d,} Diagnosis strategies for unknown (MCI and SMC) subjects. Compared to known subject recognition, identifying unknown subjects is more complicated, and more examinations are needed.\label{strategy}}
\end{figure*}

\subsection{Potential clinical applications}

OpenClinicalAI enables the AD diagnosis system to be implemented in uncertain and complex clinical settings thereby  reducing the workload of AD diagnosis and minimizing the cost to subjects.

To identify the known (AD and CN) subjects with high confidence, the operating point of OpenClinicalAI runs with a high decision threshold ($0.95$). For the test set, OpenClinicalAI achieves an accuracy value of $92.47\%$, an AD sensitivity value of $84.92\%$, and a CN sensitivity value of $81.27\%$ while retaining an unknown sensitivity value of $93.96\%$. In addition, it can cooperate with the senior clinician to identify the rest of the known subjects, which are not marked as any known kinds (AD or CN). In this work, $15.08\%$ of AD subjects and $18.73\%$ of CN subjects are marked as unknown and sent to senior clinicians for diagnosis. This is significant for undeveloped areas, since it  is a promising way to connect developed and undeveloped areas to reduce workload, improve overall medical services, and promote medical equity.
To minimize the subject cost and maximize the subject benefit, our method dynamically develops personalized diagnosis strategies for the subject according to the subject's situation and existing medical conditions.

OpenClinicalAI judges whether it can finalize the subject's diagnosis according to the currently obtained information of subjects. If the current data are insufficient to establish a high confidence diagnosis, it will provide recommendations for the most appropriate next steps. This approach effectively tackles the issue of overtesting, resulting in reduced costs for subjects while maximizing the benefits they receive. For the test set, $35$ different diagnosis strategies are applied to the subject by OpenClinicalAI (as shown in Table S2). The details of the high-cost examination are as follows:

(1) $31.07\%$ of subjects require a nuclear magnetic resonance scan.
(2) $12.72\%$ of subjects require a positron emission computed tomography scan with $18$-FDG.
(3) $8.29\%$ of subjects require a positron emission computed tomography scan with AV$45$.
(4) $8.39\%$ of subjects require a gene analysis.
(5) $4.48\%$ of subjects require a cerebral spinal fluid analysis.

For the medical institution, before the system recommends an examination for a subject, OpenClinicalAI will inquire whether the medical institution can execute the examination. Suppose the medical institution cannot perform the examination. In this case, OpenClinicalAI will recommend other examinations until the current information of the subject is enough to support it to make a diagnosis or until all common examinations have been suggested and the subject is marked as unknown. This enables OpenClinicalAI to be deployed in different medical institutions with varying examination capabilities. In this study, OpenClinicalAI conducted subject diagnoses for $40$ examination conditions that could potentially be encountered in a health care facility (Table S3). Additionally, OpenClinicalAI made $14,654$ adjustments to diagnostic strategies for subjects in the test set who lacked the necessary information to receive examination recommendations. This suggests that the medical facility may have been incapable of performing the recommended examinations.

\section{Conclusion}

After comparing the performance of state-of-the-art models for AD diagnosis in both closed clinical and real-world settings, we noticed that the models that performed exceptionally well in the closed clinical setting did not maintain the same level of effectiveness in the real-world setting. This suggests it is time to switch attention from algorithmic research in closed clinical setting settings to systematic study in real-world settings while focusing on the challenge of tackling the uncertainty and complexity of real-world settings. In this work, we have proposed a novel open, dynamic machine learning framework to allow the model to directly address uncertainty and complexity in the real-world setting. The resulting AD diagnostic system demonstrates great potential to be implemented in real-world settings with different medical environments to reduce the workload of AD diagnosis and minimize the cost to the subject.

Although many AI diagnostic systems have been proposed, how to embed these systems into the current health care system to improve medical services remains an open issue~\cite{mckinney2020international,kuo2021application,schneider2021reflections,bullock2020mapping}. 
OpenClinicalAI provides a reasonable way to embed the AI system into the current health care system. OpenClinicalAI can collaborate with clinicians to improve the clinical service quality, especially the clinical service quality of undeveloped areas. On the one hand, OpenClinicalAI can directly deal with the diagnosis task in an uncertain and complex real-world setting. On the other hand, OpenClinicalAI can diagnose typical patients of known subjects while sending those challenging or atypical patients of known subjects to clinicians for diagnosis. Although AI technology is different from traditional statistics, the model of the AI system still learns patterns from training data. The model can easily learn patterns from typical patients, but it can be challenging to learn patterns for atypical patients. Thus, every atypical and unknown patient is especially needed to be treated by  clinicians. In this work, most of the known subjects are diagnosed by OpenClinicalAI, and the rest are marked as unknown and sent to the senior clinician.

Although OpenClinicalAI is promising for impacting future research on the diagnosis system, several limitations remain. First, prospective clinical studies of the diagnosis of Alzheimer's disease will be required to prove the effectiveness of our system. Second, the data collection and processing are required to follow the standards of the ADNI.

% \section{Contributors}
\section{CRediT authorship contribution statement}

\textbf{Yunyou Huang: }Conceptualized, Methodology, Model design, Coding, Data curation, Writing the original draft. \textbf{Xiaoshuang Liang: }Writing review, Data curation. \textbf{Suqin Tang: }Writing review, Data curation. \textbf{Li Ma: }Writing review, Data curation. \textbf{Fan Zhang: }Data curation. \textbf{Fan Zhang: }Data curation. \textbf{Xiuxia Miao: }Data curation, Software. \textbf{Xiangjiang Lu: }Data curation, Software. \textbf{Jiyue Xie: }Data curation, Software. \textbf{Zhifei Zhang} and \textbf{Jianfeng Zhan: } Supervision, Conceptualization, Funding acquisition, Project administration, Writing review \& editing.  

\section{Declaration of competing interests}

The authors declare no competing financial interest.

% \section{Data and code availability}
\section{Data availability}

The data from the Alzheimer's Disease Neuroimaging Initiative were used under license for the current study. Applications for access to the dataset can be made at~\url{http://adni.loni.usc.edu/data-samples/access-data/}. All original code has been deposited at the website ~\url{https://www.benchcouncil.org/}{BenchCouncil} and is publicly available when this article is published.

\section{Acknowledgment}
We thank Weibo Pan and Fang Li for downloading the raw datasets from the Alzheimer's Disease Neuroimaging Initiative. This work is supported by the Standardization Research Project of Chinese Academy of Sciences (No. BZ201800001 to J.Z. ), the Project of Guangxi Science and Technology (No. GuiKeAD20297004 to Y.H. ), and the National Natural Science Foundation of China (No. 61967002 and No. U21A20474 to S.T. ).

 \bibliographystyle{elsarticle-num-names} 
 \bibliography{main}




% \end{thebibliography}
\end{CJK}
\end{document}

\begin{CJK}{UTF8}{gbsn}
\begin{flushleft}
 \textbf{\large{Supplementary Material:}}
\end{flushleft}
\begin{center}
    \textbf{\large{OpenClinicalAI: An Open and Dynamic Model for Alzheimer's Disease Diagnosis}}
\end{center}

 \noindent{Yunyou Huang, Xiaoshuang Liang, Xiangjiang Lu, Xiuxia Miao, Jiyue Xie, Wenjing Liu, Fan Zhang, Guoxin Kang, Li Ma, Suqin Tang, Zhifei Zhang$^*$, Jianfeng Zhan$^*$}\\

 \noindent{$^*$Corresponding authors’ emails: zhifeiz@ccmu.edu.cn (Zhifei Zhang), zhanjianfeng@ict.ac.cn (Jianfeng Zhan)}
~\\
%\noindent{\textbf{This file includes:}}
\renewcommand*{\contentsname}{This file includes:}
\begin{spacing}{2}
\tableofcontents
\end{spacing}

\clearpage
\renewcommand{\figurename}{Figure S}
\setcounter{figure}{0}
%\renewcommand{\thefigure}{\ifnum \c@chapter>\z@ \thechapter-\fi \@arabic\c@figure}
%\noindent\textbf{Supplemental information}
%\vskip 20pt
%\noindent\textbf{Supplementary materials}
%~\\
%\begin{figure*}[ht]

% \subsection{Figure S 2}
% \begin{figure*}[ht]
% \centering
% \includegraphics[height=12cm, width=13cm]{image/open_clinical_setting.eps} 
% \caption{\textbf{The real-world setting of Clinical AIBench.}  Subjects in real-world settings are different with various situations. They contain different pre-known categories and unknown and unfamiliar categories for the specific clinician or AI diagnostic system. The visit of subjects to a particular medical institution can not be pre-specified and hence are uncertain. Medical institutions in real-world  settings also are different with different executive abilities of the examination. The executive ability of the examination in various medical institutions is very different from small-scale country clinics to large-scale hospitals. In addition, it is difficult to know by advance all the specific medical institutions that will deploy the AI system and their particular executive abilities of the examination. \label{open_setting}}
% \end{figure*}

\clearpage
\subsection{Dataset Detail}
\label{Dataset}
\noindent Data used in the preparation of this article were obtained from the Alzheimer's Disease Neuroimaging Initiative (ADNI) database (\url{http://adni.loni.usc.edu}). The ADNI was launched in $2003$ as a public-private partnership, led by Principal Investigator Michael W. Weiner,MD. For up-to-date information, see \url{http://www.adni-info.org.}

The data is collected from $67$ sites in the United States and Canada ~\cite{petersen2010alzheimer,weiner2010alzheimer,weiner2015impact,weiner2017alzheimer}. The subject in the dataset aged between $54.4$ and $91.4$ at the first visit. The interval of the subject follow-up is usually greater than $6$ months. Generally, the longer the follow-up time is, the longer the interval is. The first visit is marked as bl, and the other visit is marked as m\underline{xx} according to the time (For example, the visit takes place six months after the first visit is marked as m$06$). 
% Detailed characteristics of the subject are shown in Table. S\ref{ptd}.
~\\
% Detailed characteristics of the subject are shown in Table S\ref{ptd},\ref{visit}.

\noindent The data contains study data, image data, genetic data compiled by ADNI between 2005 and 2019. Considering the commonly used examinations and the concerned examinations in AD diagnosis by the clinician, 13 categories of data are selected. 
\begin{itemize}
 \item[(1)]Base information, usually obtained through consultation, includes demographics, family history, medical history, symptoms. 
 \item[(2)]Cognition information, usually obtained through consultation and testing, includes Alzheimer's Disease Assessment Scale, Mini-Mental State Exam, Montreal Cognitive Assessment, Clinical Dementia Rating, Cognitive Change Index. 
 \item[(3)]Cognition testing, usually obtained through testing, includes ANART, Boston Naming Test, Category Fluency-Animals, Clock Drawing Test, Logical Memory-Immediate Recall, Logical Memory-Delayed Recall, Rey Auditory Verbal Learning Test, Trail Making Test. 
 \item[(4)]Neuropsychiatric information, usually obtained through consultation, includes Geriatric Depression Scale, Neuropsychiatric Inventory, Neuropsychiatric Inventory Questionnaire. 
 \item[(5)]Function and behavior information, usually obtained through consultation, includes Function Assessment Question, Everyday Cognitive Participant Self Report, Everyday Cognition Study Partner Report. 
 \item[(6)] Physical, neurological examination, usually obtained through testing, includes Physical Characteristics, Vitals, neurological examination. 
 \end{itemize}
 
The rest of the examinations include blood testing, urine testing, nuclear magnetic resonance scan, positron emission computed tomography scan with 18-FDG, positron emission computed tomography scan with AV45, gene analysis, and cerebral spinal fluid analysis. It is worth noting that not all categories of information are obtained for a subject's visit, and the information on each type is often incomplete.

\clearpage
\subsection{Figure S 1}
\begin{figure*}[ht]
\centering
\includegraphics[height=9.13cm, width=16cm]{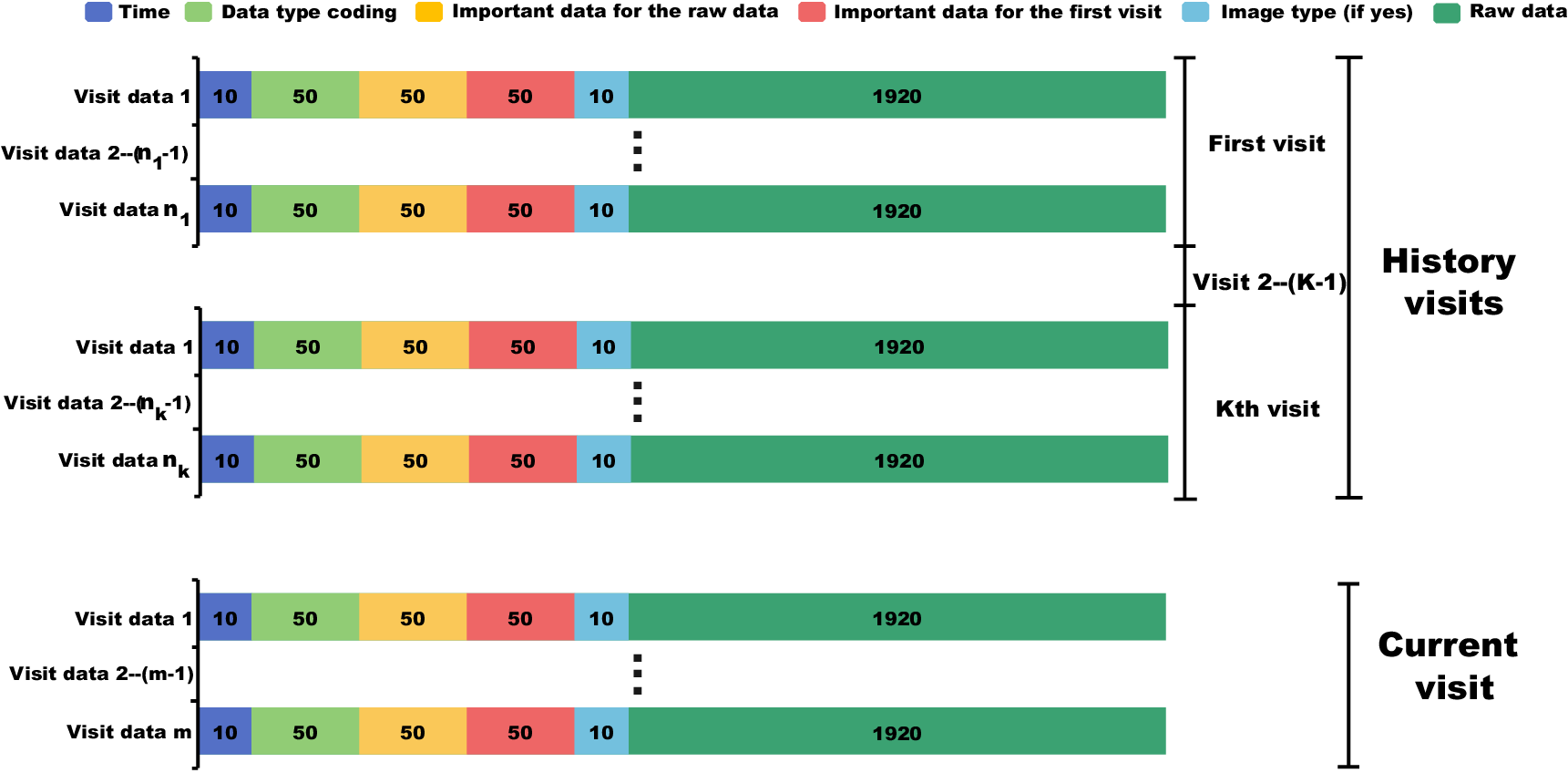} 
\caption{\textbf{Data framework for a single subject.} Our data representation framework comprehensively considers the historical visit information and current visit information of the subject. The data with the earlier time is farther away from the current data. \label{Extent_fig2}}
\end{figure*}

\clearpage
\subsection{Table S 1}
\begin{table*}[ht]
\centering
\caption{\textbf{The visit distribution of subjects.}\label{visit}}
\begin{tabular}{ccccc}
\hline
Visit & Data set & Training set & Validation set & Test set \\
\hline
first visit    & 2126     & 705          & 53             & 1368     \\
m06    & 1515     & 604          & 45             & 866      \\
m12   & 1475     & 621          & 43             & 811      \\
m18   & 329      & 79           & 5              & 245      \\
m24   & 1217     & 591          & 32             & 594      \\
m36   & 804      & 338          & 23             & 443      \\
m48   & 638      & 311          & 14             & 313      \\
m60   & 399      & 178          & 13             & 208      \\
m72   & 395      & 207          & 14             & 174      \\
m84   & 268      & 128          & 6              & 134      \\
m96   & 146      & 74           & 3              & 69       \\
m108  & 100      & 55           & 1              & 44       \\
m120  & 75       & 39           & 1              & 35       \\
m132  & 55       & 33           & 0              & 22       \\
m144  & 39       & 19           & 1              & 19       \\
m156  & 12       & 4            & 0              & 8       \\
\hline
\end{tabular}
\end{table*}

\clearpage

\subsection{Table S 2}
\begin{table*}[ht]
\centering
\tiny
\caption{\textbf{The diagnosis strategies for the test set.}\label{methond}}

\begin{tabular}{cccccccccccccccccc}
\hline
&\multicolumn{13}{c}{Diagnosis strategies}                                          & \multicolumn{4}{c}{Visit number of subject} \\
&Base & Cog & CE & Neur & FB & PE & Blood & Urine & MRI & FDG & AV45 & Gene & CSF & AD      & CN      & Unknown     & Total     \\
\hline
\textbf{1}&1    & 1   & 0  & 0    & 0  & 0  & 0     & 0     & 0   & 0   & 0    & 0    & 0   & 244     & 280     & 2680        & 3204      \\
\textbf{2}&1    & 1   & 1  & 1    & 1  & 1  & 0     & 0     & 1   & 0   & 0    & 0    & 0   & 16      & 23      & 697         & 736       \\
\textbf{3}&1    & 1   & 1  & 1    & 1  & 1  & 0     & 0     & 1   & 0   & 1    & 0    & 0   & 3       & 10      & 81          & 94        \\
\textbf{4}&1    & 1   & 1  & 1    & 1  & 1  & 0     & 0     & 1   & 1   & 1    & 0    & 0   & 6       & 8       & 124         & 138       \\
\textbf{5}&1    & 1   & 1  & 0    & 0  & 0  & 0     & 0     & 0   & 0   & 0    & 0    & 0   & 8       & 47      & 43          & 98        \\
\textbf{6}&1    & 1   & 1  & 1    & 1  & 1  & 0     & 0     & 0   & 0   & 0    & 0    & 0   & 10      & 6       & 183         & 199       \\
\textbf{7}&1    & 1   & 1  & 1    & 0  & 0  & 0     & 0     & 0   & 0   & 0    & 0    & 0   & 1       & 2       & 27          & 30        \\
\textbf{8}&1    & 1   & 1  & 1    & 1  & 1  & 0     & 0     & 0   & 1   & 0    & 0    & 0   & 1       & 2       & 11          & 14        \\
\textbf{9}&1    & 1   & 1  & 1    & 1  & 1  & 0     & 0     & 0   & 0   & 1    & 0    & 0   & 1       & 1       & 16          & 18        \\
\textbf{10}&1    & 1   & 1  & 1    & 1  & 1  & 0     & 0     & 1   & 1   & 0    & 0    & 0   & 2       & 6       & 232         & 240       \\
\textbf{11}&1    & 1   & 1  & 1    & 1  & 1  & 0     & 0     & 0   & 1   & 1    & 0    & 0   & 0       & 3       & 16          & 19        \\
\textbf{12}&1    & 1   & 1  & 1    & 1  & 0  & 0     & 0     & 0   & 1   & 1    & 0    & 0   & 0       & 0       & 1           & 1         \\
\textbf{13}&1    & 1   & 1  & 1    & 1  & 0  & 0     & 0     & 0   & 0   & 0    & 0    & 0   & 1       & 0       & 33          & 34        \\
\textbf{14}&1    & 1   & 0  & 1    & 1  & 1  & 0     & 0     & 1   & 1   & 1    & 0    & 0   & 0       & 0       & 2           & 2         \\
\textbf{15}&1    & 1   & 0  & 1    & 1  & 1  & 0     & 0     & 1   & 0   & 0    & 0    & 0   & 0       & 0       & 1           & 1         \\
\textbf{16}&1    & 1   & 1  & 1    & 1  & 1  & 1     & 0     & 1   & 1   & 1    & 1    & 1   & 2       & 0       & 113         & 115       \\
\textbf{17}&1    & 1   & 1  & 1    & 1  & 1  & 1     & 1     & 0   & 0   & 0    & 0    & 0   & 5       & 14      & 10          & 29        \\
\textbf{18}&1    & 1   & 1  & 1    & 1  & 1  & 0     & 0     & 1   & 0   & 0    & 1    & 0   & 0       & 3       & 21          & 24        \\
\textbf{19}&1    & 1   & 1  & 1    & 1  & 1  & 1     & 0     & 1   & 1   & 1    & 1    & 0   & 3       & 0       & 13          & 16        \\
\textbf{20}&1    & 1   & 1  & 1    & 1  & 1  & 1     & 0     & 0   & 0   & 0    & 0    & 0   & 1       & 0       & 42          & 43        \\
\textbf{21}&1    & 1   & 1  & 1    & 1  & 1  & 1     & 0     & 1   & 0   & 0    & 0    & 0   & 0       & 1       & 2           & 3         \\
\textbf{22}&1    & 1   & 1  & 1    & 1  & 1  & 0     & 0     & 1   & 0   & 1    & 1    & 0   & 0       & 5       & 21          & 26        \\
\textbf{23}&1    & 1   & 1  & 1    & 1  & 1  & 0     & 0     & 1   & 1   & 0    & 1    & 0   & 1       & 0       & 9           & 10        \\
\textbf{24}&1    & 1   & 1  & 1    & 1  & 1  & 1     & 1     & 1   & 1   & 0    & 1    & 1   & 0       & 0       & 54          & 54        \\
\textbf{25}&1    & 1   & 1  & 1    & 1  & 1  & 1     & 0     & 1   & 0   & 0    & 1    & 0   & 0       & 0       & 60          & 60        \\
\textbf{26}&1    & 1   & 1  & 1    & 1  & 1  & 1     & 1     & 1   & 0   & 0    & 1    & 1   & 0       & 0       & 67          & 67        \\
\textbf{27}&1    & 1   & 1  & 1    & 1  & 1  & 1     & 0     & 1   & 1   & 0    & 1    & 0   & 0       & 0       & 55          & 55        \\
\textbf{28}&1    & 1   & 1  & 1    & 1  & 1  & 1     & 1     & 1   & 1   & 0    & 1    & 0   & 0       & 0       & 1           & 1         \\
\textbf{29}&1    & 1   & 1  & 1    & 1  & 1  & 1     & 1     & 1   & 0   & 0    & 1    & 0   & 0       & 0       & 3           & 3         \\
\textbf{30}&1    & 1   & 1  & 1    & 1  & 1  & 1     & 0     & 1   & 1   & 0    & 1    & 1   & 0       & 0       & 2           & 2         \\
\textbf{31}&1    & 1   & 1  & 1    & 1  & 1  & 0     & 0     & 1   & 1   & 1    & 1    & 0   & 0       & 0       & 13          & 13        \\
\textbf{32}&1    & 1   & 1  & 1    & 1  & 1  & 1     & 0     & 1   & 1   & 1    & 0    & 0   & 0       & 0       & 1           & 1         \\
\textbf{33}&1    & 1   & 1  & 1    & 1  & 1  & 1     & 0     & 1   & 0   & 0    & 1    & 1   & 0       & 0       & 1           & 1         \\
\textbf{34}&1    & 1   & 1  & 1    & 1  & 1  & 0     & 0     & 1   & 0   & 0    & 1    & 1   & 0       & 0       & 1           & 1         \\
\textbf{35}&1    & 1   & 1  & 1    & 1  & 1  & 0     & 0     & 0   & 0   & 1    & 1    & 0   & 0       & 0       & 1           & 1        \\
\hline
\end{tabular}
\end{table*}

\clearpage
\subsection{Table S 3}
\begin{table*}[ht]
\centering
\tiny
\caption{\textbf{Medical institutions with different examination abilities in the test set.} \label{institutions}}
\begin{threeparttable}
\begin{tabular}{cccccccccccccccccc}
\hline
&\multicolumn{13}{c}{\begin{tabular}[c]{@{}c@{}}Medical institution \\without examination capabilities$^1$\end{tabular}} & \multicolumn{4}{c}{\begin{tabular}[c]{@{}c@{}}Visit number of subject in the \\ condition of medical institution\end{tabular}} \\
&Base       & Cog       & CE       & Neur       & FB       & PE       & Blood       & Urine       & MRI       & FDG       & AV45      & Gene      & CSF      & AD                           & CN                          & Unknown                          & Total                          \\
\hline
\textbf{1}&0          & 0         & 0        & 0          & 0        & 0        & 1           & 1           & 0         & 0         & 0         & 0         & 0        & 23                           & 35                          & 952                              & 1010                           \\
\textbf{2}&0          & 0         & 0        & 0          & 0        & 0        & 1           & 1           & 0         & 1         & 0         & 1         & 1        & 0                            & 2                           & 11                               & 13                             \\
\textbf{3}&0          & 0         & 0        & 0          & 0        & 0        & 1           & 1           & 0         & 1         & 0         & 0         & 0        & 3                            & 8                           & 63                               & 74                             \\
\textbf{4}&0          & 0         & 0        & 0          & 0        & 0        & 1           & 1           & 1         & 0         & 0         & 0         & 0        & 1                            & 4                           & 23                               & 28                             \\
\textbf{5}&0          & 0         & 0        & 0          & 0        & 0        & 1           & 1           & 1         & 1         & 0         & 1         & 1        & 1                            & 0                           & 1                                & 2                              \\
\textbf{6}&0          & 0         & 0        & 0          & 0        & 0        & 1           & 1           & 0         & 1         & 1         & 1         & 1        & 0                            & 2                           & 39                               & 41                             \\
\textbf{7}&0          & 0         & 0        & 0          & 0        & 0        & 1           & 1           & 1         & 0         & 0         & 1         & 1        & 0                            & 1                           & 4                                & 5                              \\
\textbf{8}&0          & 0         & 0        & 0          & 0        & 0        & 0           & 0           & 0         & 0         & 1         & 0         & 0        & 3                            & 0                           & 19                               & 22                             \\
\textbf{9}&0          & 0         & 0        & 0          & 0        & 0        & 0           & 0           & 1         & 0         & 0         & 0         & 0        & 1                            & 0                           & 3                                & 4                              \\
\textbf{10}&0          & 0         & 0        & 0          & 0        & 0        & 1           & 1           & 1         & 1         & 1         & 1         & 1        & 1                            & 1                           & 11                               & 13                             \\
\textbf{11}&0          & 0         & 0        & 0          & 0        & 0        & 1           & 1           & 1         & 1         & 0         & 0         & 0        & 0                            & 1                           & 13                               & 14                             \\
\textbf{12}&0          & 0         & 0        & 0          & 0        & 0        & 1           & 1           & 0         & 0         & 1         & 1         & 1        & 0                            & 0                           & 26                               & 26                             \\
\textbf{13}&0          & 0         & 0        & 0          & 0        & 1        & 1           & 1           & 1         & 0         & 0         & 0         & 0        & 0                            & 0                           & 1                                & 1                              \\
\textbf{14}&0          & 0         & 0        & 0          & 0        & 0        & 0           & 0           & 0         & 1         & 1         & 0         & 0        & 0                            & 0                           & 69                               & 69                             \\
\textbf{15}&0          & 0         & 0        & 0          & 0        & 0        & 0           & 0           & 1         & 1         & 0         & 0         & 0        & 0                            & 0                           & 1                                & 1                              \\
\textbf{16}&0          & 0         & 0        & 0          & 0        & 0        & 1           & 1           & 0         & 0         & 0         & 1         & 1        & 0                            & 0                           & 19                               & 19                             \\
\textbf{17}&0          & 0         & 0        & 0          & 0        & 0        & 1           & 1           & 0         & 0         & 1         & 0         & 0        & 0                            & 1                           & 6                                & 7                              \\
\textbf{18}&0          & 0         & 1        & 0          & 0        & 0        & 1           & 1           & 0         & 0         & 0         & 1         & 1        & 0                            & 0                           & 1                                & 1                              \\
\textbf{19}&0          & 0         & 1        & 0          & 0        & 0        & 1           & 1           & 0         & 0         & 0         & 0         & 0        & 0                            & 0                           & 2                                & 2                              \\
\textbf{20}&0          & 0         & 0        & 0          & 0        & 0        & 1           & 1           & 0         & 0         & 1         & 1         & 0        & 0                            & 0                           & 5                                & 5                              \\
\textbf{21}&0          & 0         & 0        & 0          & 0        & 0        & 0           & 1           & 0         & 0         & 0         & 0         & 0        & 12                           & 45                          & 85                               & 142                            \\
\textbf{22}&0          & 0         & 0        & 0          & 0        & 0        & 1           & 1           & 0         & 1         & 1         & 0         & 1        & 0                            & 3                           & 21                               & 24                             \\
\textbf{23}&0          & 0         & 0        & 0          & 0        & 0        & 0           & 0           & 0         & 0         & 0         & 1         & 0        & 1                            & 0                           & 0                                & 1                              \\
\textbf{24}&0          & 0         & 0        & 0          & 0        & 0        & 1           & 1           & 0         & 1         & 0         & 0         & 1        & 0                            & 4                           & 10                               & 14                             \\
\textbf{25}&0          & 0         & 0        & 0          & 0        & 0        & 1           & 1           & 0         & 0         & 1         & 0         & 1        & 1                            & 0                           & 9                                & 10                             \\
\textbf{26}&0          & 0         & 0        & 0          & 0        & 0        & 0           & 1           & 0         & 1         & 1         & 0         & 1        & 0                            & 0                           & 60                               & 60                             \\
\textbf{27}&0          & 0         & 0        & 0          & 0        & 0        & 0           & 1           & 0         & 0         & 1         & 0         & 1        & 0                            & 0                           & 54                               & 54                             \\
\textbf{28}&0          & 0         & 0        & 0          & 0        & 0        & 0           & 0           & 0         & 0         & 1         & 0         & 1        & 0                            & 0                           & 1                                & 1                              \\
\textbf{29}&0          & 0         & 0        & 0          & 0        & 0        & 0           & 0           & 0         & 1         & 1         & 0         & 1        & 0                            & 0                           & 3                                & 3                              \\
\textbf{30}&0          & 0         & 0        & 0          & 0        & 0        & 0           & 1           & 0         & 0         & 0         & 0         & 1        & 0                            & 0                           & 5                                & 5                              \\
\textbf{31}&0          & 0         & 0        & 0          & 0        & 0        & 1           & 1           & 0         & 0         & 0         & 0         & 1        & 0                            & 0                           & 9                                & 9                              \\
\textbf{32}&0          & 0         & 0        & 0          & 0        & 0        & 0           & 1           & 0         & 1         & 1         & 0         & 0        & 0                            & 0                           & 1                                & 1                              \\
\textbf{33}&0          & 0         & 0        & 0          & 0        & 0        & 1           & 1           & 0         & 1         & 1         & 0         & 0        & 0                            & 0                           & 1                                & 1                              \\
\textbf{34}&0          & 0         & 0        & 0          & 0        & 0        & 1           & 1           & 0         & 1         & 1         & 1         & 0        & 0                            & 0                           & 2                                & 2                              \\
\textbf{35}&0          & 0         & 0        & 0          & 0        & 0        & 0           & 0           & 0         & 0         & 1         & 1         & 0        & 0                            & 0                           & 5                                & 5                              \\
\textbf{36}&0          & 0         & 0        & 0          & 0        & 0        & 0           & 0           & 0         & 1         & 0         & 0         & 0        & 0                            & 0                           & 1                                & 1                              \\
\textbf{37}&0          & 0         & 0        & 0          & 0        & 0        & 0           & 1           & 0         & 0         & 1         & 0         & 0        & 0                            & 0                           & 1                                & 1                              \\
\textbf{38}&0          & 0         & 0        & 0          & 0        & 0        & 1           & 0           & 0         & 1         & 0         & 0         & 0        & 0                            & 0                           & 1                                & 1                              \\
\textbf{39}&0          & 0         & 0        & 0          & 0        & 0        & 1           & 1           & 1         & 1         & 0         & 0         & 1        & 0                            & 0                           & 1                                & 1                              \\
\textbf{40}&0          & 0         & 0        & 0          & 0        & 0        & 1           & 0           & 0         & 1         & 1         & 0         & 0        & 0                            & 0                           & 1                                & 1  \\
\hline                           
\end{tabular}
\begin{tablenotes}
        \footnotesize
        \item[1] The examination is marked as 1, meaning that the medical institution cannot perform this examination for the subject. The examination is marked as 0, indicating that (1) the medical institution can perform this examination for the subject, or (2)  OpenClinicalAI does not request for performing this examination during the diagnosis of the subject though the medical institution may not be able to perform this examination for the subject. It is worth noting that the examination ability in the test set may be different from other AI systems since 0 may mean that OpenClinicalAI does not request for performing this examination during the diagnosis of the subject. However, the medical institution may not be  able to perform this examination for the subject.
  \end{tablenotes}
\end{threeparttable}
\end{table*}
\clearpage

\subsection{Table S 4}
\begin{table*}[ht]
\centering
\caption{\textbf{SNPs relate to AD.}\label{snp_ad}}
\begin{tabular}{cccc}
\hline
SNP\_NAME   & SNP\_NAME   & SNP\_NAME   & SNP\_NAME   \\
\hline
rs429358    & rs7412      & rs10948363  & rs7274581   \\
rs17125944  & rs4147929   & rs6656401   & rs11771145  \\
rs6733839   & rs983392    & rs10498633  & rs28834970  \\
rs9271192   & rs35349669  & rs9331896   & rs1476679   \\
rs10792832  & rs2718058   & rs190982    & rs10838725  \\
rs11218343  & rs4844610   & rs10933431  & rs9271058   \\
rs75932628  & rs9473117   & rs12539172  & rs10808026  \\
rs73223431  & rs3740688   & rs7933202   & rs3851179   \\
rs17125924  & rs12881735  & rs3752246   & rs6024870   \\
rs7920721   & rs138190086 & rs4723711   & rs4266886   \\
rs61822977  & rs6733839   & rs10202748  & rs115124923 \\
rs115675626 & rs1109581   & rs17265593  & rs2597283   \\
rs1476679   & rs78571833  & rs12679874  & rs2741342   \\
rs7831810   & rs1532277   & rs9331888   & rs7920721   \\
rs3740688   & rs7116190   & rs526904    & rs543293    \\
rs11218343  & rs6572869   & rs12590273  & rs7145100   \\
rs74615166  & rs2526378   & rs117481827 & rs7408475   \\
rs3752246   & rs7274581   &             &            \\
\hline
\end{tabular}
\end{table*}

\clearpage

% \usepackage{multirow}
\subsection{Table S 5}
\begin{table*}[ht]
\centering
\caption{\textbf{The normal range of indicators.}\label{index_nomal}}
\begin{threeparttable}
\begin{tabular}{cccccc}
\hline
                                                                                                        &                                                                       & \multicolumn{2}{c}{AD\_Normal} & \multicolumn{2}{c}{CN\_Normal} \\
                                                                                                        &                                                                       & Low            & High          & Low            & High          \\
 \hline
\multirow{2}{*}{\begin{tabular}[c]{@{}c@{}}Medical \\ history\end{tabular}}                             & Psychiatric                                                           & 0              & 0             & 0              & 0             \\
                                                                                                        & \begin{tabular}[c]{@{}c@{}}Neurologic \\ (other than AD)\end{tabular} & 0              & 0             & 0              & 0             \\
\multirow{2}{*}{Symptoms\tnote{1} }                                                                               & Present\_count\_21\tnote{2}                                                      & 0              & 6             & 0              & 6             \\
                                                                                                        & Present\_count\_28\tnote{3}                                                      & 0              & 8             & 0              & 8             \\
\multirow{2}{*}{\begin{tabular}[c]{@{}c@{}}Cognitive \\ Change Index\tnote{4} \end{tabular}}                      & Score\_12\tnote{5}                                                               & 32.2188        & 60            & 12             & 13.5634       \\
                                                                                                        & Score\_20\tnote{6}                                                               & 50.3438        & 100           & 20             & 22.0845       \\
\multicolumn{2}{c}{CDRSB\tnote{7} }                                                                                                                                                       & 2              & 18            & 0              & 0             \\
\multirow{3}{*}{\begin{tabular}[c]{@{}c@{}}Alzheimer's Disease \\ Assessment Scale\tnote{8} \end{tabular}}        & ADAS11\tnote{9}                                                                & 10             & 70            & 0              & 11.264        \\
                                                                                                        & ADAS13\tnote{10}                                                                & 18             & 85            & 0              & 17.67         \\
                                                                                                        & ADASQ4                                                                & 5              & 10            & 0              & 6             \\
\multicolumn{2}{c}{MMSE\tnote{11}  }                                                                                                                                                        & 0              & 27            & 25             & 30            \\
\multicolumn{2}{c}{MOCA\tnote{12}  }                                                                                                                                                        & 0              & 23            & 26             & 30            \\
\multirow{2}{*}{\begin{tabular}[c]{@{}c@{}}Preclinical Alzheimer's \\ Cognitive Composite\tnote{13}  \end{tabular}} & mPACCdigit                                                            & -30.0745       & -7.6955       & -5.1733        & 4.7304        \\
                                                                                                        & mPACCtrailsB                                                          & -29.7277       & -6.7798       & -4.8523        & 4.3338     \\
 \hline                                                                                              
\end{tabular}
\begin{tablenotes}
        \footnotesize
        \item[1] Nausea, Vomiting, Diarrhea, Constipation, Abdominal discomfort, Sweating, Dizziness, Low energy, Drowsiness, Blurred vision, Headache, Dry mouth, Shortness of breath, Coughing, Palpitations, Chest pain, Urinary discomfort (e.g., burning), Urinary frequency, Ankle swelling, Musculoskeletal pain, Rash, Insomnia, Depressed mood, Crying, Elevated mood, Wandering, Fall, Other.
        \item[2] Nausea to Rash
        \item[3] Nausea to Other
        \item[4] The CCI scale is in \url{https://adni.bitbucket.io/reference/cci.html}.
        \item[5] CCI1 to CCI12
        \item[6] CCI1 to CCI20
        \item[7] The CDR scale is in \url{https://adni.bitbucket.io/reference/cdr.html}.
        \item[8] The Alzheimer's Disease Assessment Scale-Cognitive scale is in \url{https://adni.bitbucket.io/reference/adas.html}.
        \item[9]  Q1 to Q11
        \item[10] Q1 to Q13
        \item[11] The Mini-Mental State Exam scale is in \url{https://adni.bitbucket.io/reference/mmse.html}.
        \item[12] The Montreal Cognitive Assessment scale is in \url{https://adni.bitbucket.io/reference/moca.html}.
        \item[13] The calculation method of Preclinical Alzheimer's Cognitive Composite  is in \url{https://ida.loni.usc.edu/pages/access/studyData.jsp?categoryId=16&subCategoryId=43}.
  \end{tablenotes}
\end{threeparttable}
\end{table*}

% \clearpage

%\renewcommand{\algorithmname}{Extended Data Algorithm}
% \floatname{algorithm}{Algorithm S}
% \setcounter{algorithm}{0}
% \subsection{Algorithm S 1}
% \begin{algorithm*}[ht]
% %\scriptsize
%          \renewcommand{\algorithmicrequire}{\textbf{Input:}}
%          \renewcommand{\algorithmicensure}{\textbf{Output:}}
%          \caption{\textbf{The examination label algorithm.}\label{alg1}}
%          %\label{alg1}
%          \begin{algorithmic}[1]
%                    \REQUIRE The label set $y_{true}$, the prediction set $y_{pred}$, diagnosis strategy set $exam\_strategy$ for a subject in a visit.
%                    \ENSURE Next examination set $next\_exam$
%                    \STATE Sort the $exam\_strategy$ by the number of examinations in a diagnosis strategy.
%                    \FOR{$ i=0  \quad to  \quad len($exam\_strategy$) $}
%                    \FOR{$ j=i+1  \quad to  \quad len($exam\_strategy$) $}
%                    \IF{$exam\_strategy$[i]$\subset$$exam\_strategy$[j]}
%                     \STATE  $gain=sum(y_{true}[j]\times y_{pred}[j]-y_{true}[i]\times y_{pred}[i])+sum(\sim y_{true}[i]\times y_{pred}[i]-\sim y_{true}[j]\times y_{pred}[j])$
%                 \IF{$gain>0$}
%                 \STATE  The next examination of current examination strategy $exam\_strategy$[i] is label by $exam\_strategy$[j].
%                 \ENDIF
%                     \ENDIF
%                    \ENDFOR
%                    \ENDFOR
%          \end{algorithmic}
% \end{algorithm*}

\clearpage
\subsection{Algorithm S 1}
\begin{algorithm*}[ht]
\centering
%\scriptsize
         \renewcommand{\algorithmicrequire}{\textbf{Input:}}
         \renewcommand{\algorithmicensure}{\textbf{Output:}}
         \caption{\textbf{The modified OpenMax algorithm.}\label{alg2}}
         %\label{alg1}
         \begin{algorithmic}[1]
                   \REQUIRE The abnormal pattern dataset $X$, the FitHigh function from libMR~\citep{Scheirer_2011_TPAMI}, the MiniBatchKMeans function from scikit-learn~\citep{sculley2010web}, the number of the center of known categories of subject $N$, quantiles $Q$.
                   \ENSURE The centers of known categories of subject $C$, and libMR models $Model$, the threshold of known categories of subject $Thr$.
                   \STATE $X[i]$ is the abnormal pattern dataset of $ith$ known categories of subject, in which every data $x\in X$ is belong to $ith$ known categories of subject and is correctly classified by the trained  AI model. $L$ is the number of the known categories of subject.
                   \FOR{$ i=0 \quad to  \quad (L-1)$}
                   \STATE $C[i]=MiniBatchKMeans(X[i],N[i])$
                   \ENDFOR
                   \STATE $Dist=[]$
                   \FOR{$ i=0 \quad to  \quad (L-1)$}
                   \FOR{$ x \quad in  \quad X[i]$}
                   \STATE 
                   $Dist[i]$.add(distance(x,$C[i]$,$C_{others}$) \quad // $distance=sqrt(min\_distance(x,C[i])^2+(1-min\_distance(x,C_{others}))^2)$
                   \ENDFOR
                   \ENDFOR
                   \FOR{$ i=0 \quad to  \quad (L-1)$}
                   \STATE $Model[i]$=FitHigh($Dist[i]$)
                   \STATE $Thr[i]$ is the $Q[i]$ quantile of the $Dist[i]$
                   \ENDFOR
                   \STATE Return $C$, $Model$, $Thr$
         \end{algorithmic}
\end{algorithm*}

\clearpage

\subsection{Algorithm S 2}
\begin{algorithm*}[ht]
%\scriptsize
         \renewcommand{\algorithmicrequire}{\textbf{Input:}}
         \renewcommand{\algorithmicensure}{\textbf{Output:}}
         \caption{\textbf{OpenMax probability estimation.}\label{alg3}}
         %\label{alg1}
         \begin{algorithmic}[1]
                   \REQUIRE Abnormal pattern of the subject $X=\{x_1, x_2, ..., x_n\}$, raw data of subject $Z$, activation vector $V(Z)=\{v_1(Z), v_2(Z)\}$, The centers of known categories of subject $C$, and libMR models $Model$, the threshold of known categories of subject $Thr$, flag $F$, the numer of “top” classes to revise $\alpha$.
                   \ENSURE The prediction probability $\hat{P}$.
                   \STATE $L$ is the number of the known categories of subject.
                   \STATE Let $s(i)=argsort(v_j(Z))$
                   \STATE Let $Dist=[]$
                   \FOR{$ i=0\quad to  \quad (L-1)$}
                   \STATE $dist[i]=distance(X,C[i],C_{others})$
                   \ENDFOR
                    \FOR{$ i=1\quad to  \quad \alpha$}
                    %\STATE $\omega_{s(i)}(y)=1-\frac{2-i}{2}e^{-(\frac{\parallel y-\tau_{s(i)}\parallel}{\lambda_{s(i)}})^{\kappa_{s(i)}}}$
                   \STATE $\omega_{i}(Z)=1-\frac{\alpha-i}{\alpha}*models[i-1].w\_score($dist$[i-1])$
                   \ENDFOR
                   \STATE Revise activation vector $\hat{V}(Z) = V(Z)\circ\omega(Z)$
                   \STATE Define $\hat{v_0}(Z)=\sum_iv_i(Z)(1-\omega_i(Z))$
                   \STATE $\hat{P}(y=j\mid Z)=\frac{e^{\hat{v}_j(Z)}}{\sum_{i=0}^2e^{\hat{v}_i(Z)}}$
                   \IF{$F$}
                    \STATE $abnor\_score=[]$
                    \FOR {$ j=1\quad to  \quad (L-1)$}
                    \STATE $diff=dist[j-1]-thr[j-1]$
                    \IF{$diff<=0$}
                    \STATE  $abnor\_score$.append(0)
                    \ELSE
                     \STATE $tmp\_abnor\_score=diff/thr[j-1]$
                     \IF{$tmp\_abnor\_score>1$}
                     \STATE $tmp\_abnor\_score=1$
                     \ENDIF
                      \STATE $abnor\_score$.append($tmp\_abnor\_score$)
                    \ENDIF
                    \ENDFOR
                    \FOR {$ j=1\quad to  \quad (L-1)$}
                     \STATE $\hat{P}(y=j\mid Z)=\hat{P}(y=j\mid Z)*(1-abnor\_score[j-1])$
                    \ENDFOR
                     \STATE$\hat{P}(y=0\mid Z)=1-\sum_{j=1}^{L-1}\hat{P}(y=j\mid Z)$
                    \ENDIF
                   \STATE Return $\hat{P}$
         \end{algorithmic}
\end{algorithm*}

\clearpage
\vskip 12pt
\bibliography{main}
\end{CJK}